\documentclass[sigconf]{acmart}
\AtBeginDocument{%
  }

\setcopyright{acmlicensed}

\usepackage{mdframed,lipsum}
\usepackage{graphicx}
\usepackage{color}

\newcommand{\stitle}[1]{\vspace{1ex} \noindent{\bf #1}}
\newcommand{\nop}[1]{}
\usepackage{textcomp}
\usepackage{bm}
\usepackage{float}
\usepackage{xcolor}
\usepackage{color, colortbl}
\usepackage{url}
\usepackage{multirow}
\usepackage[ruled,linesnumbered, vlined]{algorithm2e}

\usepackage{subcaption}
\usepackage{mathtools}
\usepackage{pifont,dsfont}
\usepackage{threeparttable}
\usepackage{booktabs}
\usepackage{enumitem}
\usepackage{subfiles}

\def\A{{\bf A}}

\def\W{{\bf W}}

\def\Z{{\bf Z}}

\def\0{{\bf 0}}
\def\1{{\bf 1}}

\newcommand{\system}{\textsf{CeFGC}}
\newcommand{\advanced}{\textsf{CeFGC$^*$}}

\usepackage{xcolor,colortbl}

\copyrightyear{2026}
\acmYear{2026}
\setcopyright{cc}
\setcctype{by}
\acmConference[KDD '26]{Proceedings of the 32nd ACM SIGKDD Conference on Knowledge Discovery and Data Mining V.1}{August 09--13, 2026}{Jeju Island, Republic of Korea}
\acmBooktitle{Proceedings of the 32nd ACM SIGKDD Conference on Knowledge Discovery and Data Mining V.1 (KDD '26), August 09--13, 2026, Jeju Island, Republic of Korea}
\acmPrice{}
\acmDOI{10.1145/3770854.3780262}
\acmISBN{979-8-4007-2258-5/2026/08}

\begin{document}

\title{Communication-efficient Federated Graph Classification via Generative Diffusion Modeling}
\author{Xiuling Wang}
\email{xiulingwang@hkbu.edu.hk}
\affiliation{%
  \institution{Hong Kong Baptist University}
  \city{Hong Kong}
  \country{China}
}
\author{Xin Huang}
\email{xinhuang@comp.hkbu.edu.hk}
\affiliation{%
  \institution{Hong Kong Baptist University}
    \city{Hong Kong}
    \country{China}
}
\author{Haibo Hu}
\email{haibo.hu@polyu.edu.hk}
\affiliation{%
  \institution{Hong Kong Polytechnic University}
    \city{Hong Kong}
    \country{China}
}
\author{Jianliang Xu}
\email{xujl@comp.hkbu.edu.hk}
\affiliation{%
  \institution{Hong Kong Baptist University}
    \city{Hong Kong}
    \country{China}
}

\begin{abstract}

Graph Neural Networks (GNNs) unlock new ways of learning from graph-structured data, proving highly effective in capturing complex relationships and patterns. Federated GNNs (FGNNs) have emerged as a prominent distributed learning paradigm for training GNNs over decentralized data. However, FGNNs face two significant challenges: high communication overhead from \emph{multiple rounds} of parameter exchanges and \emph{non-IID data characteristics} across clients. To address these issues, we introduce \system, a novel FGNN paradigm that facilitates efficient GNN training over non-IID data by limiting communication between the server and clients to three rounds only.
The core idea of \system\ is to leverage generative diffusion models to minimize direct client-server communication. Each client trains a generative diffusion model that captures its local graph distribution and shares this model with the server, which then redistributes it back to all clients. Using these generative models, clients generate synthetic graphs combined with their local graphs to train local GNN models. Finally, clients upload their model weights to the server for aggregation into a global GNN model.
We theoretically analyze the I/O complexity of communication volume to show that \system\ reduces to \emph{a constant of three communication rounds only}. 
Extensive experiments on several real graph datasets demonstrate the effectiveness and efficiency of \system\ against state-of-the-art competitors, reflecting our superior performance on non-IID graphs by aligning local and global model objectives and enriching the training set with diverse graphs. 
\end{abstract}


\maketitle

\vspace{-0.05in}
\section{Introduction}
\label{sec:intro}
\begin{figure*}[t!]
    \centering
\begin{subfigure}[b]{.97\textwidth}
      \centering
\includegraphics[width=\textwidth]{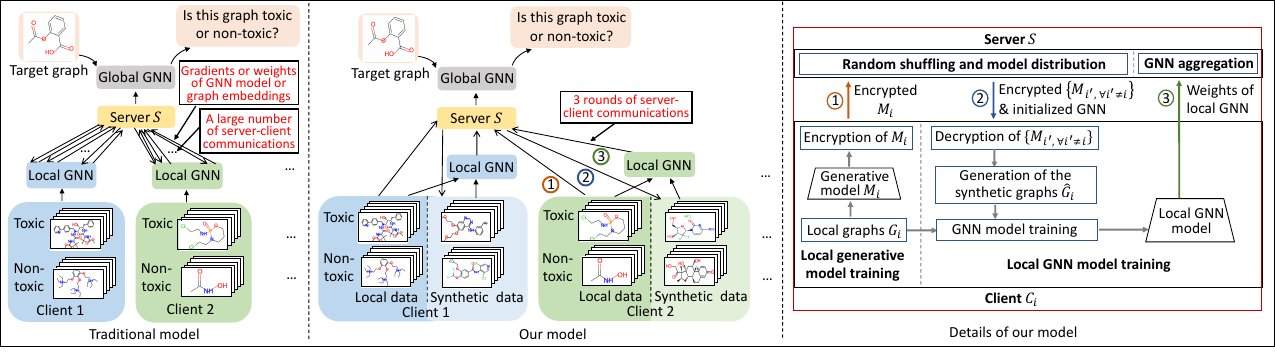}
    \end{subfigure}
    \vspace{-0.2in}
\caption{\label{fig:comp} Comparison of the framework of traditional FGNN and our model. The arrows with the numbers \ding{182}, \ding{183}, and \ding{184} indicate the communication between the server and the clients.} 
 \vspace{-0.1in}
\end{figure*}

In recent years, the ubiquity and complexity of graph-structured data have driven the advancement of Graph Neural Networks (GNNs) \cite{wu2020comprehensive, zhou2020graph} across a wide range of fields, e.g., recommender systems \cite{gao2023survey, gao2022graph}, social network analysis \cite{fan2019graph, min2021stgsn}, and biology network analysis \cite{jin2021application, zhang2021graph}. Owing to privacy concerns, commercial competition, and regulatory restrictions, Federated Learning (FL) \cite{he2020fedml, fu2022federated, liu2022federated} is designed to enable collaborative learning over decentralized data.
FL enables distributed model training among multiple clients while safeguarding local data privacy. 
There have been considerable research efforts in Federated Graph Neural Networks (FGNNs) \cite{he2020fedml, fu2022federated, liu2022federated}, which can be categorized into two types \cite{liu2022federated}: {\em horizontal FGNNs}, where clients' data share the same node features but possess different node samples, and {\em vertical FGNNs}, where clients' local data exhibit different node features. This paper focuses on horizontal FGNNs, which typically involve a central server and a set of clients, each possessing a set of graphs. Each client independently trains a local GNN on its own data, then sends model parameters/gradients or node/graph embeddings to the server, which aggregates these inputs to produce the global GNN \cite{he2021fedgraphnn, fu2022federated, liu2022federated}. In this paper, we focus on graph classification \cite{kudo2004application, wu2020comprehensive} as a global GNN's learning task, which aims to classify unlabeled graphs into different categories. 
Figure \ref{fig:comp} illustrates a toy example of graph classification using FGNNs that clients collaboratively train a global GNN to predict whether a given test graph is toxic or non-toxic 
by server–client exchanges.

\vspace{0.01in}
\noindent{\bf Motivations and challenges.} While FGNNs show promise in learning over decentralized graph data, they encounter two major challenges. First, the intensive server-client updates during training cause significant communication overhead \cite{mcmahan2017communication, li2020federated, kairouz2021advances}, especially for large models, bandwidth-limited or offline clients \cite{konevcny2016federated, huang2013depth}, as synchronous processing is often costly. Therefore, we focus on reducing communication overhead by enabling asynchronous processing via diffusion models. Second, the data distribution on different clients may diverge significantly on node features and 
graph structures \cite{xie2021federated}, which introduces drift in clients' local updates, thereby hindering model convergence and accuracy \cite{gao2022feddc,wu2021fedgnn}.

To address these challenges, we design \system, a novel framework for graph classification over non-IID graphs that requires only three rounds of server-client communications. {Here, we define a server-client communication round as a single exchange of data between the server and clients, in either direction.}
Briefly, \system\ is a four-step process. 
First, each client trains a graph generative diffusion model (GGDM) \cite{cao2022survey, yang2023diffusion} on its local graphs, which captures the underlying distribution of these graphs. The clients then share these GGDMs with the server (the first round of communication).
The server, in turn, redistributes GGDMs and an initialized GNN to clients (the second round of communication).  
Armed with the received GGDMs, each client generates a set of synthetic graphs and trains a local GNN model using both these synthetic graphs and its local graphs.
Subsequently, the clients transmit the weights of their local GNNs to the server (the third round of communication). The server aggregates these weights to obtain the global model.
By leveraging GGDMs, \system\ achieves two key benefits. First, \system\ not only minimizes the server-client communication rounds but also reduces server-side computation to a single aggregation. Second, it better aligns local objectives with the global model, while also enhancing model's generalization capability by enriching each client's training set with more diverse graphs.
These advantages significantly enhance the efficiency and effectiveness of graph classification over non-IID graphs.

Two primary challenges remain for the design of \system. First, since previous GGDMs do not incorporate graph labels into the training process, each client must train one diffusion model per class. This results in high communication bandwidth when transmitting GGDMs in the system, 
especially with a large number of data classes and clients. The key challenge is how to reduce the number of GGDMs per client while still capturing class-conditional distributions. Second, sharing GGDMs with an honest-but-curious server or other clients poses potential privacy risks, potentially exposing sensitive information of clients' local graphs. 
Specifically, \system\ faces two potential privacy risks: (i) the server may attempt to infer the clients' local data from the uploaded GGDMs; (ii) the clients may try to extract sensitive information about other clients' local data from the distributed GGDMs. 

To tackle the first challenge, we introduce a label channel into the training process of GGDM, which takes the label of the graph as input and outputs a noise vector associated with that label. This enables different classes of data to be trained with varying levels of noise within a single diffusion model, allowing each client to train just one model instead of multiple models.

For the second challenge, we have the following solutions. First, each client shares an encrypted GGDM with the server and provides the public key to other clients, ensuring the server cannot infer local data from the uploaded models. Second, the server randomly shuffles the GGDMs before distribution, preventing clients from identifying models' origins. Furthermore, we propose a privacy-enhanced method where the server aggregates each diffusion model with the most similar one before distribution, further mitigating privacy leakage risks.
Additionally, it is important to note that common inference attacks \cite{wang2019beyond, hitaj2017deep, shokri2017membership, melis2019exploiting, zhu2019deep, zhao2020idlg} or poisoning attacks \cite{yin2021comprehensive, bouacida2021vulnerabilities, lyu2020threats, mothukuri2021survey} against FL models are ineffective in our model, as these attacks typically rely on large numbers of communication rounds, such as analyzing gradient or embedding differences between rounds, which is not applicable in our \system. 

\noindent{\bf Our contributions.} To the best of our knowledge, this is the first work to design communication-efficient FGNNs for non-IID graph data. In summary, our model introduces two major advancements: first, our model requires only three rounds of server-client communications, efficiently addressing the challenge of communication overhead in traditional FGNNs; and second, by introducing GGDMs to FGNN, each client trains a local GNN based on both the local and synthetic data, and thus effectively improves performance on non-IID graphs across clients. Figure \ref{fig:comp} illustrates the key differences between traditional FGNNs and our proposed model. We summarize our main contributions below. 
\begin{itemize}[leftmargin=*]
\item 
We introduce \system, a framework that leverages both generative diffusion models and a novel model distribution scheme to minimize the communication overhead of FGNNs while maintaining high model accuracy over non-IID data.
\item We modify the existing graph diffusion generative model by incorporating a label channel into the training process, 
which reduces the communication bandwidth required by the system.
\item We conduct comprehensive experiments on three non-IID settings with five real-world datasets from different domains. Our results show that, despite the minimal involvement of only three rounds of server-client communication, \system\ consistently outperforms the SOTA baselines across different settings. 
\end{itemize}

\vspace{-0.1in}
\section{Related Work}
\label{sec:related work}
\stitle{FGNNs with non-IID data for graph classification.} 
Recently, several works investigated how to deal with non-IID data with FGNNs.
GCFL \cite{xie2021federated} is a graph-clustered FL framework that clusters clients based on their data heterogeneity, where clients with less data heterogeneity are grouped together. FedStar \cite{tan2023federated} 
captures and shares the universal structural knowledge across multiple clients using a particular type of feature-structure decoupled GNN \cite{dwivedi2021graph}, thereby enhancing the local performance of clients in FGNN. GCFGAE \cite{guo2023globally} integrates FGNN with split learning by dividing the model into portions trained independently by clients and the server. However, GCFGAE focuses on unsupervised learning and is not applicable to our setting. Notably, all the above works face communication overhead, requiring numerous server-client interactions.

\stitle{Communication efficiency in FL.}
Numerous works have been introduced to reduce the communication overhead in FL \cite{shahid2021communication, konevcny2016federated, chen2021communication, sattler2019robust}, including parameter compression \cite{chen2021communication, rothchild2020fetchsgd, reisizadeh2020fedpaq}, client selection \cite{abdulrahman2020fedmccs, xu2020client, cho2020client}, reducing model updates \cite{kamp2019efficient, nguyen2020fast, wu2021fast}, and one-shot FL \cite{zhou2020distilled, guha2019one}. However, most of these works focus on learning over non-graph data, leaving the reduction of communication overhead in FGNNs largely unexplored. 
In summary, the issue of communication overhead in FedGNNs remains unsolved.

\stitle{Generative diffusion models for graphs.}
Generative diffusion models 
have become an emerging generative paradigm for graph generation. The goal of graph generative diffusion models is to learn the underlying distribution of the given graphs and generate novel graphs.  
The existing studies 
can be broadly categorized into three classes \cite{zhang2023survey, fan2023generative}: (1) {\em Score-based Generative Models (SGM)} \cite{niu2020permutation, chen2022nvdiff} that employ a score function to illustrate the probability distribution of the data; 
(2) {\em Denoising Diffusion Probabilistic Models (DDPMs)} \cite{vignac2022digress, haefeli2022diffusion} that add discrete Gaussian noise to the graph with Markov transition kernels \cite{austin2021structured} and trains a neural network to predict the added noise to recover the original graph, and 
(3) {\em Stochastic Differential Equations (SDEs)} \cite{huang2022graphgdp, jo2022score, luo2022fast} that characterize the development of a system over time under the effect of random noise. 
We refer the readers to some surveys on generative diffusion models on graphs 
\cite{fan2023generative,zhang2023survey}. 
This paper uses the pioneer EDP-GNN \cite{niu2020permutation} as the generative diffusion model. 

\vspace{-0.05in}
\section{Preliminaries}
\label{sc:pre}
\subsection{Graph Neural Network} 
Consider a graph $G(V, E, X)$ with node set $V$, edge set $E$, and node features $X$, a Graph Neural Network (GNN) aims to learn representations of $G$ by aggregating information from a node’s neighbors using neural networks. The learned embedding can be applied to various graph analytics tasks, such as node classification and graph classification. In this paper, we focus on graph classification, which aims to determine the label of an entire graph (e.g., a molecule's toxicity). For this task, all node embeddings will be transformed into a single graph embedding to determine the graph's label.

Most of the existing GNNs follow the message-passing neural network (MPNN) \cite{wu2020comprehensive} to learn the node embedding. It starts by initializing the node embedding with node features. Each node then receives and aggregates "messages" from its neighbors to form intermediate embeddings. After 
$k$ steps, each node embedding incorporates information from its $k$‐hop neighbors. 
These node embeddings are then aggregated by a {\em graph pooling operation}, such as max, mean, or hierarchical pooling \cite{hamilton2020graph, ying2018hierarchical}, to form a whole graph embedding \cite{ying2018hierarchical, zhang2018end}. 
In this paper, we use GIN \cite{xu2018powerful} as the GNN model and mean pooling for aggregation. 

\vspace{-0.07in}
\subsection{Fedrated Graph Neural Networks}
A Federated Graph Neural Network (FGNN) consists of a server $S$ and a set of $N$ clients $\mathcal{C}=\{C_i\}_{i=1}^{N}$. Each client $ C_i$ owns a set of private local graphs $\mathcal{G}_i$ sampled from its own data distribution. 
$S$ and $C$ will jointly learn a global GNN model over $\mathbb{G}=\bigcup_{i=1}^{N}\{\mathcal{G}_i\}$. The objective function of the global GNN model is to optimize the overall objectives while keeping private data locally \cite{fu2022federated}:
\vspace{-0.05in}
\begin{equation}
\small
\label{eqn:obj-FGNN}
\mathop{min}\limits_{\W}\sum_{i=1}^{N} \mathcal{L}_{i}(\mathcal{G}_i;\W),
\end{equation} 
where \W\ represents the parameters of the global GNN, and $\mathcal{L}_{i}(\mathcal{G}_i;\W)$ is the loss (such as cross-entropy loss) over the local graphs $\mathcal{G}_i$ on client $C_i$. As we consider graph classification as the downstream task, the output of the global GNN is the graph embedding $\Z_i$ for each graph $G_i\in\mathbb{G}$. We assume there are $K$ unique classes within $\mathbb{G}$, denoted as $Y=\{1,...,K\}$. Each graph embedding $\Z_i$ is then used to assign the corresponding graph a label within $Y$.

\vspace{-0.07in}
\subsection{Generative Diffusion Models on Graphs}

Graph generation models aim to generate new graph samples that resemble a given dataset. 
Among these, diffusion-based models have become increasingly popular, which gradually introduce noise into data until it conforms to a prior distribution \cite{niu2020permutation, huang2022graphgdp, chen2022nvdiff, jo2022score, vignac2022digress, haefeli2022diffusion, luo2022fast}. 

Generally, existing graph generative diffusion models (GGDMs) include two processes: (1) the \textit{forward process}, which progressively degrades the original data into Gaussian noise \cite{ho2020denoising}, and (2) the \textit{reverse process}, which gradually denoises the noisy data back to its original structure using transition kernels.
In this paper, we use Score-based Generative Modeling (SGM) for graph generation. Specifically, given a probability density function $p(x)$, SGM aims to estimate the data score function $\nabla_x log p_{data}(x)$ in the forward process. It perturbs the data with Gaussian noise of varying intensities and jointly estimates the scores for all noise levels. For each noise level $\sigma$, it trains a noise conditional model $s_\theta(x;\sigma)$ to approximate its score function. Then, given a noise distribution $q_{\sigma}(\hat{x}|x)$, where $\hat{x}$ is a noisy version of $x$, and a set of noise levels $\{\sigma_l\}^L_{l=1}$, where $L$ is the number of levels, the training loss is defined as: 
\vspace{-0.05in}
\begin{equation}
\small
\label{eqn:obj-diffusoin0}
\begin{aligned}
\mathcal{L}(\theta; \{\sigma_l\}^L_{l=1})=
\mathop{min}\limits_{\theta}\frac{1}{2L}\sum_{l=1}^{L} {\sigma_l^2} \mathbb{E}\left[\left\| s_{\theta}(\hat{x}_l, \sigma_l) -  \nabla_{\hat{x}_l}\log q_{\sigma_l}(\hat{x}_l|x) \right\|_2^2\right],
\end{aligned}
\end{equation} 
where $\mathbb{E}$ is the expectation.

In the reverse process, after obtaining the trained conditional score model $s_\theta(x;\sigma)$, synthetic graphs are generated using noise-conditional score networks, such as the Score Matching with Langevin Dynamics model \cite{song2019generative}. It is worth noting that although we use SGM-based model in this paper, 
any other GGDMs that can capture the graph distribution can be adapted to our framework.
\vspace{-0.05in}
\section{Details of \system}
\label{sc:model}
In this section, we formalize the problem and present  our \system. 

\vspace{-0.05in}
\subsection{Problem Formulation}
Given a set of labeled graphs $\mathbb{G}=\bigcup_{i=1}^{N}\{\mathcal{G}_i\}$, a server $S$ and a set of $N$ clients $\mathcal{C}=\{C_i\}_{i=1}^{N}$. Each client $C_i$ holds a set of private graphs $\mathcal{G}_i$ sampled from $\mathbb{G}$, with each set following its own data distribution. The objective of our model is to collaboratively train a global GNN, denoted as $GNN_{global}$, that can perform graph classification, i.e., predicting the label $y_{target}$ of a given graph $G_{target}$, based on all the $\mathcal{G}_i$ from the clients. This problem can be formalized as follows:
\begin{equation}
\small
\label{eqn:katt}
GNN_{global}: \{\mathbb{G}, \mathcal{C}, GNN_{initial}, G_{target} \} \rightarrow y_{target},
\end{equation}
where $GNN_{initial}$ represents the initialization of the GNN model. 

Next, we present \system, which incorporates generative models into FGNNs to address the server-client communication overhead and improve the model performance over non-IID graphs.
Overall, \system\ includes four steps: (1) training of the local generative model, (2) transferring of local generative models among server and clients, (3) training of local GNN models, and (4) generation of the global GNN. 
The rightmost figure in Figure \ref{fig:comp} illustrates the steps of \system.   
Next, we explain the details of each step.

\vspace{-0.05in}
\subsection{Phase 1: Local Generative Model Training}
We propose two methods for training local generative models: (1) the {\em basic method}, where each client trains a set of EDP-GNN \cite{niu2020permutation}, with each associated to a specific graph class; and (2) the {\em advanced method}, where each client trains a single generative model over all local graphs using EDP-GNN equipped with a class label channel.

\subsubsection{Basic method}

For any client $C_i$, let $\mathcal{G}_i$ be the set of local graphs owned by $C_i$. First, $C_i$ groups all local graphs in $\mathcal{G}_i$ by their class labels, assigning graphs with the same label to the same group. Let $\mathcal{G}_i^k$ be the set of graphs in $\mathcal{G}_i$ that are associated with the label $y_k$. 
Next, $C_i$ trains a set of generative models $\mathcal{M}_i=\{M_{i,k}\}$, each $M_{i,k}$ is trained on all graphs in group $\mathcal{G}_i^k$. 
In this paper, we adapt EDP-GNN \cite{niu2020permutation}, one of the SGM models for graph generation, to our setting. 
EDP-GNN trains a set of conditional score functions for different levels of Gaussian noise.

Specifically, given a graph $G$ with its adjacency matrix $\A$, $\A$ is perturbed with the Gaussian noise of the distribution $q_{\sigma}(\hat{\A}|\A)$:
\vspace{-0.07in}
\begin{equation}
\small
\label{eqn:noise}
    q_{\sigma}(\hat{\A}|\A) = \prod_{i<j}\frac{1}{\sqrt{2\pi\sigma}}\exp{-\frac{(\hat{\A}_{i,j}-{{\A}_{i,j}}^2)}{2\sigma^2}}, 
\end{equation} 
where $\hat{\A}$ represents the perturbed adjacency matrix, and $\sigma$ is the standard deviation of Gaussian noise. EDP-GNN varies the $\sigma$ values and obtains a set of noise graphs at different noise intensity levels $\{\sigma_{l}, \forall l \in [1, L]\}$, where $L$ is the total number of the noise level. 

Given the original $\A$ and perturbed adjacency matrices  $\hat{\A}=\{\hat{\A}_l, \forall l \in [1, L]\}$ corresponding to the noise set $\{\sigma_{l}, \forall l \in [1, L]\}$, the objective function of EDP-GNN is to minimize the training loss of the conditional score model $s_{\theta}(\hat{\A}, \theta)$: 
\vspace{-0.05in}
\begin{equation}
\label{eqn:obj-diffusoin}
\mathop{min}\limits_{\theta}\frac{1}{2L}\sum_{l=1}^{L} {\sigma_l^2} \mathbb{E}\left[\left\| s_{\theta}(\hat{\A}_l, \sigma_l) - \frac{\hat{\A}_l-\A}{\sigma_l^2} \right\|_2^2\right],
\end{equation} 
where $\mathbb{E}$ is the expectation. 
This objective function can be optimized by an existing optimizer such as SGD \cite{ruder2016overview} and Adam \cite{kingma2014adam}. 

In this method, $C_i$ trains a set of generative models $\mathcal{M}_i=\{M_{i,k}\}$ for all graph groups. The number of generative models in $\mathcal{M}_i$ is equal to the number of distinct labels in $\mathcal{G}_i$. Assuming there are $N$ clients participating in FL training, the dataset comprises $K$ classes, and each diffusion model requires $s$ storage space, the communication bandwidth in terms of the diffusion model for {\em Basic method} is $N \times K \times s$. Consequently, the communication bandwidth can become substantial with a large number of clients, or numerous classes. Therefore, to reduce the communication bandwidth, we propose an advanced method in which each client $C_i$ only needs to train one generative model. This modification reduces the communication bandwidth to $N \times s'$, with $s \approx s'$. The details of the advanced method are as follows.

\subsubsection{Advanced method}
The key innovation of the {\em Advanced method} is incorporating a class label channel during the training of the generative model. This involves adding an embedding layer that takes the data class label as input and outputs a vector of different noise levels for each class label. The noise vector is subsequently employed in both the diffusion forward and reverse processes.
By employing this approach, different classes of graphs can be trained with varying noise levels. Specifically, the class label-based embedding layer can be represented as follows:
\vspace{-0.05in}
\begin{equation}
\label{eqn:sigma}
{\boldsymbol{\sigma}}_{k}=(\sigma_{1,k},\sigma_{2,k},..., \sigma_{L,k})=emb(k),
\end{equation} 
where $k$ represents one of the class labels in the dataset, $emb$ denotes the embedding layer, and $L$ is the pre-defined noise level.
Then, the training loss of the diffusion model in Eq. \ref{eqn:obj-diffusoin0} can be modified as:
\vspace{-0.05in}
\begin{equation}
\label{eqn:obj-diffusoin0-modified}
\begin{aligned}
\small
&\mathcal{L}(\theta'; \{\sigma_{l, k}\}^{L, K}_{l=1, k=1})=
\mathop{min}\limits_{\theta'}\frac{1}{2*L*K}\sum_{k=1}^{K} \sum_{l=1}^{L} {\sigma_{l, k}^2} \cdot\\
&\mathbb{E}\left[ \mathds{1}_{y(x)=k} \left\| s_{\theta'}(\hat{x}_{l,k}, \sigma_{l,k}) -  \nabla_{\hat{x}_{l,k}}\log q_{\sigma_{l,k}}(\hat{x}_{l,k}|x) \right\|_2^2\right],
\end{aligned}
\end{equation} 
where $\theta'$ denotes the parameters of the generative model, $K$ represents the number of class in the dataset, and $y(x)$ indicates the class label of data record $x$. Accordingly, Formula \ref{eqn:obj-diffusoin} can be rewritten as:
\vspace{-0.05in}
\begin{equation}
\label{eqn:obj-diffusoin-modified}
\begin{aligned}
\small
\mathop{min}\limits_{\theta'}\frac{1}{2*L*K}\sum_{k=1}^{K} \sum_{l=1}^{L}{\sigma_{l,k}^2} \cdot
\mathbb{E}\left[ \mathds{1}_{y(\A)=k} \left\| s_{\theta'}(\hat{\A}_{l,k}, \sigma_{l,k}) - \frac{\hat{\A}_{l,k}-\A}{\sigma_{l,k}^2} \right\|_2^2\right],
\end{aligned}
\end{equation} 
where $y(\A)$ is the class label of the graph with adjacency matrix $\A$. 

\vspace{-0.05in}
\subsection{Phase 2: Generative Model Transferring}
After each client $C_i$ obtains his/her locally trained generative models $\mathcal{M}_i$, he/she uploads $\mathcal{M}_i$ to the server $S$. To prevent privacy leakage from each client, we require all clients to encrypt their diffusion models using the same encryption mechanism. Subsequently, all the encrypted diffusion models are uploaded to the server, ensuring that the server cannot infer the data information from the clients (the details of our privacy-enhancing method are provided later in Section \ref{sc:enhanced}). The server then combines the generative models from all clients, excluding the model from the client itself. After that, the server applies a random shuffle, and distributes the packaged generative models back to the clients, along with the initialization of the GNN. This approach prevents clients from associating the generative models with their respective IDs.

It is worth noting that first, clients are not required to possess samples from all classes. Second, sending generative models instead of generated graphs to server reduces communication bandwidth as well as storage requirements for both server and clients.

\begin{table}[t!]
    \centering
\scalebox{0.73}{\begin{tabular}{|c|c|c|c|c|c|}
    \hline
         {\textbf{Settings}}&{\textbf{Round 1}}&\textbf{Round 2}&\textbf{Round 3}&\textbf{> Round 3}&\textbf{I/O complexity}\\
         \hline
         $\textbf{Traditional}$&\multicolumn{4}{c|}{\multirow{2}{*}{$D(GNN)*N/round$}}&{\multirow{2}{*}{$D(GNN)*N*S$}}\\
         $\textbf{FGNNs}$&\multicolumn{4}{c|}{ }&\\\hline
         $\textbf{\advanced}$&$D(DM)$&$D(DM)$&$D(GNN)$&\multirow{2}{*}{-}&$D(DM)*N^2+$\\
         $\textbf{\system}$&$*N$&$*(N-1)*N$&$*N$&&$D(GNN)*N$\\
         \hline
    \end{tabular}}
    \caption{ Theoretical analysis of communication volume.}
    \label{tab:volume}
    \vspace{-0.2in}
\end{table}
\vspace{-0.05in}

\subsection{Phase 3: Local GNN Model Training}

For each client $C_i$, let $\mathcal{M}^S_i$ be the generative models $C_i$ receives and decrypts from the server. First, $C_i$ generates a set of synthetic graphs $\hat{\mathcal{G}}_i$, each corresponding to a generative model $M\in\mathcal{M}^S_i$. 
For each $M$, $C_i$ can generate multiple graphs that follow the same distribution modeled by $M$. 
Next, $C_i$ trains a local GNN model on the training data consisting of all the generated synthetic graphs as well as $C_i$'s local graphs (i.e., $\hat{\mathcal{G}}_i \cup \mathcal{G}_i$).
Finally, $C_i$ uploads the parameters $\W_i$ of the local GNN model to the server. 

For generating the synthetic graphs, we adopt the annealed Langevin dynamics sampling algorithm \cite{song2019generative}. First, the client randomly samples an integer $n$, where $n$ is the number of nodes to be generated. Then, it initializes an adjacency matrix $\tilde{\A}^0 \in \mathbb{R}^{n\times n}$ by using the folded norm distribution:
\vspace{-0.05in}
\begin{equation}
\label{eqn:A_initial}
\tilde{\A}^0_{i,j}=|\varepsilon_{i,j}|, \varepsilon_{i,j} \sim \mathcal N(0,1).
\end{equation} 
Then $\Tilde{\A}$ is updated by iterative sampling from a series of trained conditional score models $\{s_{\theta}(\tilde{\A}, \sigma_l)\}_0^L$, which are the generative models $C_i$ receive and decrypt from the server.

So far, the score-based generative model can provide the synthetic sample in the continuous space, whereas the entries in the graph's adjacency matrix are discrete and binary (0/1). To obtain a proper adjacency matrix, we transform the continuous adjacency matrix $\tilde{\A}$ to a binary one:
$\tilde{\A}^{final}_{i,j} = \mathbf{1}_{\tilde{\A}_{i,j}>0.5}$. 

As the local GNNs are trained over both synthetic and local graphs that may have different distributions, these models potentially may not be able to converge. However, our empirical studies show that all local GNNs can converge well under different settings. 

\vspace{-0.05in}
\subsection{Phase 4: Global GNN Model Generation}
After collecting all local GNN models from the clients, the server employs an aggregation mechanism to combine them. In this paper, we follow FedAvg \cite{mcmahan2017communication} and aggregate model weights to construct the global model. Specifically, the weights of the global GNN at $l$-th layer are computed as follows:
\vspace{-0.15in}
\begin{equation}
\label{eqn:fedavg}
\W^{l}_{global}=\frac{1}{N}\sum_{i=1}^{N} \W^{l}_{i},
\end{equation} 
where $N$ represents the number of clients, and $\W^{l}_{i}$ denotes the weight matrix of local GNN trained by client $C_i$ at $l$-th layer. Unlike conventional FGNNs, the global GNN in \system\ is generated through a single aggregation operation rather than multiple iterations of parameter updates. This streamlined approach enhances the efficiency and simplicity of \system. Despite utilizing a single aggregation computation, our empirical results demonstrate its ability to achieve superior performance in graph classification.

\begin{algorithm}[t!]
    \caption{\label{alg:model}
    Generative diffusion model based distributed graph classification}
    \KwIn{Graphs $\mathcal{G}_i, i \in \{1,2, ..., N\}$ for each Client $i$, and the total number of clients is $N$.}
    \KwOut{Global model parameters $\W_{global}$}
    \For{$i=1,2,...,N$ in parallel for all the clients}
    {
    Train the generative models $M_{i}$ for $\mathcal{G}_i$ based on the {\em basic method} or {\em advanced method}\;
    Encrypt $M_{i}$ and upload the encrypted $M_{i}$ to the central server\;
    }
    The central server collects all the generative models from the clients, randomly shuffles and aggregates (optional) the generative models, and sends them to the clients. Meanwhile, the central server sends the initialized global model parameters $\W_{global}$ to each client\;
    \For{$i=1,2,...,N$ in parallel for all the clients}
    {Decrypt the generative models from the server\;
    Generate a set of synthetic graphs $\hat{\mathcal{G}}_i$, each corresponding to a generative model $M\in\mathcal{M}^S_i$
    \;   
    Train a local GNN model on the training data consisting of all the generated synthetic graphs as well as $C_i$'s local graphs (i.e., $\hat{\mathcal{G}}_i \cup \mathcal{G}_i$)\;
    Upload the local GNN model parameters $\W_i$ to the  central server\;
    }
    The central server aggregates all the local GNN model parameters $\W_{i}$ to construct the global GNN model $\W_{global}$ by using FedAvg (Eq. \ref{eqn:fedavg})\;
    Return $\W_{global}$
\end{algorithm}
\stitle{Complexity analysis and model artifacts}.
We refer to 
our {\em Basic method} as \system\ and {\em Advanced method} as \advanced\ in the following sections.
We analyze the efficiency of our models in contrast to the traditional FGNNs such as GCFL+~\cite{xie2021federated},  FedStar~\cite{tan2023federated}, in terms of I/O complexity in Table \ref{tab:volume}. 
Let $D()$ denote data size, $DM$ the diffusion model, $GNN$ the used GNN model, $N$ the number of clients, and $S$ the total communication rounds of traditional FGNNs.
The I/O costs of \system\ and \advanced\ per round are $D(DM)*N$, $D(DM)*(N-1)*N$, $D(GNN)*N$ for the 1st, 2nd, and 3rd round, respectively. 
Therefore, the overall communication I/O complexity for both \system\ and \advanced\ is
$O(D(DM)*N+ D(DM)*(N-1)*N +D(GNN)*N)$ $= O(D(DM)*N^2 +D(GNN)*N)$.
On the contrary, for traditional FGNNs, the model needs a much larger number of rounds $S \gg 3$ case by case, as shown later in Table~\ref{tab:rounds}. Each round transmits $D(GNN)*N$ volume. Thus, the I/O complexity is $O(D(GNN)*N*S)$, where a large $S \gg 3$ leads to a lower efficiency in practical applications shown in Table~\ref{tab:cost1}. 
The pseudo-code of our model can be found in Algorithm \ref{alg:model}
\footnote{Our code is available at https://gitfront.io/r/username/5xhoUzcHcPH5/CeFGC/.}.

\vspace{-0.05in}
\section{Empirical Evaluation}
\label{sec:eval}
 \begin{table*}[t!]
    \centering
    \small
    \scalebox{0.87}{\begin{tabular}{|c|c|c|c|c|c|c|c|c|c|c|c|c|c|}
    \hline
         \multirow{3}{*}{{\bf Method}}&\multicolumn{5}{c|}{{\bf Single-dataset setting}}&\multicolumn{4}{c|}{{\bf Across-dataset setting}}  &\multicolumn{4}{c|}{{\bf Across-domain setting}}\\\cline{2-14}
          &\multirow{2}{*}{MUTAG}&\multirow{2}{*}{ENZYMES}&\multirow{2}{*}{PROTEINS}&\multirow{2}{*}{IMDB-B}&\multirow{2}{*}{IMDB-M}&\multicolumn{2}{c|}{Protein}&\multicolumn{2}{c|}{Social}  &\multicolumn{4}{c|}{Molecule\&Protein\&Social} \\\cline{7-14}
         &&&&&& ENZYMES &PROTEINS &IMDB-B &IMDB-M &MUTAG &ENZYMES &IMDB-B &IMDB-M\\\hline
          {\textbf {FedAvg}}&0.78&0.55&0.83&0.80&0.75&0.73&\bf 0.88&0.72&0.77&0.76&0.71&0.77&0.69\\
         {\textbf {FedProx}}&0.87&0.62&0.83&0.80&0.75&0.70&0.87&0.72&0.75&0.77&0.76&0.69&0.70\\{{\textbf{FedDC}}}&0.78&0.63&0.68&0.74&0.72&0.73&0.86&0.72&0.74&0.79&0.72&0.72&0.68\\{{\textbf{MOON}}}&0.86&0.63&0.71&0.79&0.70&0.72&0.87&0.71&0.76&0.77&0.71&0.73&0.70\\
         {\textbf {GCFL+}}&0.72&0.54&0.75&0.66&0.58&0.71&\bf 0.88&0.70&0.77&0.79&0.70&0.71&0.65\\
         {\textbf {FedStar}}&0.72&0.56&0.72&0.57&0.55&\bf 0.78&0.69&0.70&0.64&\bf 0.90&0.77&0.66&\bf 0.80\\
         {\textbf{One-shot}}&0.72&0.49&0.76&0.54&0.52&-&-&-&-&-&-&-&-\\\hline
         {\textbf {\system}}&\bf 0.91&\bf 0.68&\bf 0.87&\bf 0.81&\bf 0.76 &0.75&\bf 0.89&\bf 0.80&\bf 0.79&0.88&\bf 0.80&\bf 0.80&\bf 0.80\\
        {\textbf {\advanced}} &\bf 0.90&\bf 0.69&\bf 0.87&\bf 0.85&\bf 0.77&\bf 0.79&0.86&\bf 0.79&\bf 0.78&\bf 0.92&\bf 0.79&\bf 0.78&\bf 0.80\\\hline
    \end{tabular}}
\caption{Comparison of global model AUC performance. 
}
    \label{tab:performance mix auc}
   \vspace{-0.2in}
\end{table*}
We conduct a set of empirical studies, aiming to address three key research questions: (1) $\mathbf{Q_1}$ - How effective are \system\ and \advanced\ on real-world datasets which are non-IID? $\mathbf{Q_2}$ - What are the server-client communication costs of the models? (3) $\mathbf{Q_3}$ - How do the various factors affect the model performance?  
All the algorithms are implemented in Python with PyTorch and executed on NVIDIA A100-PCIE-40GB.

\vspace{-0.1in}
\subsection{Experimental Setup}
\label{sc:setup}

\stitle{Datasets.} We use five real-world datasets from three domains: a molecule (MUTAG), two proteins (ENZYMES, PROTEINS), and two social networks (IMDB-BINARY, IMDB-MULTI), each consisting of a set of graphs. These are benchmarks for graph classification \cite{morris2020tudataset}. The statistics of these datasets can be found in Appendix \ref{appx:data}. In the following text, we refer to IMDB-BINARY and IMDB-MULTI as IMDB-B and IMDB-M, respectively. 
Each dataset uses an 8:2 global train/test split. The global training set is then evenly distributed among three clients, and each client further splits the assigned graphs into train/validate/test sets with a ratio of 0.7/0.1/0.2.

We study three scenarios of distributing data over clients:
{\bf (1) Single-dataset} setting: the graphs from a single dataset are randomly distributed across the clients; 
{\bf (2) Across-dataset} setting: multi datasets from the same domain participate in the global model training, and each client holds the graphs sampled from one dataset (e.g., IMDB-B and IMDB-M datasets); 
{\bf (3) Across-domain} setting: multi datasets from different domains participate in the global model training, and each client has the graphs sampled from one dataset (e.g., MUTAG and IMDB-B datasets). 

We measure the average heterogeneity of features and structures for each setting following \cite{xie2021federated}, with details and analysis provided in Appendix \ref{appx:data}. Our observations indicate significant heterogeneity across graphs in different settings, confirming their non-IID.

\stitle{Baselines}. 
We consider two types of baselines, SOTA FL models\footnote{Implementations available at https://github.com/Oxfordblue7/GCFL and https://github.com/yuetan031/FedStar.} and one-shot FL-based method: 
\textit{\textbf{ (1) SOTA FL models:}} {\bf FedAvg} \cite{mcmahan2017communication}, {\bf FedProx} \cite{li2020federated}, {\bf GCFL+} \cite{xie2021federated}, {\bf FedStar} \cite{tan2023federated}, {{\bf FedDC} \cite{gao2022feddc} and {\bf MOON} \cite{li2021model}.} 
Both FedAvg and FedProx are representative of general FL. FedProx extended FedAvg to address the non-IID issue by adding a proximal term to reduce model differences between local and global models. GCFL+ and FedStar are SOTA FGNNs specifically designed for graph classification with non-IID data. {FedDC and MOON are two SOTA FL methods originally developed for non-IID image data, which we have adapted to our graph learning setting.} 
\textit{\textbf{(2) One-shot FL-based method:}} we adapt the one-shot model \cite{guha2019one} to our setting, where the server collects all generative models from clients without transferring them back. Instead, the server generates a set of synthetic graphs and trains a global GNN on its own.

\stitle{Evaluation metrics and parameters.} We measure the performance of GNN on graph classification{\em Accuracy} and {\em AUC}. 
{\bf (1) Generative models}: We build a 4-layer EDP-GNN\footnote{Implementation available at https://github.com/ermongroup/GraphScoreMatching} as the GGDM with the parameter setup in \cite{niu2020permutation}.
We use a range of noise scales $\sigma=\{0.1, 0.2, 0.4, 0.6, 0.8, 1.6\}$ for \system\ and set the number of noise scales to 6 for \advanced. 
{\bf (2) GNN models}: 
We use the same 3-layer GIN (hidden size 32) for all methods (except FedStar) and follow the configuration in \cite{tan2023federated} for FedStar.

\vspace{-0.05in}
\subsection{Model Performance}

In this part of the experiments, we evaluate the graph classification accuracy and AUC of the global model. 

Before presenting the results of \system\ and \advanced, we first evaluate the performance of generative diffusion models, both with and without the graph label channel. We evaluate the quality of diffusion models by employing maximum mean discrepancy (MMD) to compare the distributions of graph statistics \cite{zhang2023survey, niu2020permutation, chen2022nvdiff, huang2022graphgdp, jo2022score, luo2022fast}, 
which are commonly used to assess the model's performance. The results are available in Appendix \ref{sec:results_diffusion}. We observe comparable performance of \system\ and \advanced, suggesting that the diffusion model maintains good quality when the graph label channel is included. Next, we proceed to evaluate the model performance.

{\bf Single-dataset setting.} 
Table \ref{tab:performance mix auc} ("Single-dataset setting" column) presents the graph classification AUC of global model over global testing data for \system, \advanced, and baselines. 
We observe that both \system\ and \advanced\ consistently achieve high AUC across datasets.
For instance, the AUC on MUTAG is 0.91 with \system\ and 0.90 with \advanced. The improvements over the baselines range from 0.01 to 0.31. 
These results highlight the effectiveness of \system\ and \advanced\ in the single-dataset setting. Our improvements stem from two aspects: first, the GGDMs can effectively capture the distribution of graphs, allowing the GNN objectives for each local client to align more closely with the global model; and second, the GGDMs enhance the training set with more diverse graphs, increasing the model's generalization capability.
Furthermore, an interesting observation is that the One-shot learning-based model performs poorly in most settings, suggesting that the original graphs play a crucial role in global model training. Therefore, we omit this baseline from the subsequent evaluations.

{\bf Across-dataset setting.} Table \ref{tab:performance mix auc} (“Across-dataset setting” column) reports the AUC performance for two across-dataset settings, 
one with two protein datasets and another with two social networks.
We have several observations. First, despite higher heterogeneity in the across-dataset setting, the global model's AUC remains comparable to the single-dataset setting. For instance, when two social networks are distributed, the AUC of IMDB-M dataset is 0.79 with \system\ and 0.78 with \advanced.
Second, \system\ and \advanced consistently outperform the six baselines across all settings. For example, \system\ achieves an increment as high as 14.2\% on IMDB-B and 2.6\% on IMDB-M compared with GCFL+, while \advanced\ gains 12.9\% on IMDB-B and 1.3\% on IMDB-M.
Third, the performance for some datasets improves in the across-dataset setting compared to the single-dataset setting. For example, when the datasets are from proteins domain, 
ENZYMES and PROTEINS AUCs rise by 14.5\% and 2.3\% over single-dataset, which suggests mutual benefits from cross-dataset collaboration.

\begin{table}
{
    \centering
    \scalebox{0.83}{\begin{tabular}{|c|c|c|c|c|c|c|}
    \hline
{\multirow{2}{*}{\textbf{Method}}}&\multicolumn{3}{c|}{{\bf Communication rounds}} &\multicolumn{3}{c|}{{\bf Communication volume (Mbit)}} \\\cline{2-7}
&{MUTAG} &{Social}&{Mix} &{MUTAG} &{Social}&{Mix}\\\hline
{\textbf{FedAvg}}&392&392&192&280.00&715.35&700.75\\
{\textbf{FedProx}}&416&548&288&297.14&1,000.03&1,051.12\\ 
${\textbf{FedDC}}$&375&321&170&267.86&585.786&620.456\\
${\textbf{MOON}}$&353&301&198&252.14&549.28&722.65\\
{\textbf{GCFL+}}&712&200&164&508.57&364.97&598.56\\
{\textbf{FedStar}}&360&272&226&560.72&955.10&1,587.15\\\hline
 \textbf{\system}&\multicolumn{3}{c|}{\multirow{2}{*}{3}}&13.90&163.80&537.73\\
 \textbf{\advanced}&\multicolumn{3}{c|}{ }&\bf 7.31&\bf 69.32&\bf 191.29\\\hline
    \end{tabular}}}
    \caption{The number of communication rounds and the communication volume between server and clients under single dataset (MUTAG), across-dataset (Social), and across-domain (Mix) settings. }
   \label{tab:rounds}
   \vspace{-0.3in}
\end{table}

\begin{table*}[t!]
    \centering
    \scalebox{0.84}{\begin{tabular}{|c|c|c|c|c|c|c|c|c|c|c|c|c|}
    \hline
    \multirow{3}{*}{\textbf{Settings}}&\multicolumn{4}{c|}{Single dataset (MUTAG)}&\multicolumn{4}{c|}{Across-dataset (Social)}&\multicolumn{4}{c|}{Across-domain}\\\cline{2-13}
         &\multicolumn{2}{c|}{\textbf{LAN (45 Mb/s)}}&\multicolumn{2}{c|}{\textbf{WAN (1 Mb/s)}}&\multicolumn{2}{c|}{\textbf{LAN (45 Mb/s)}}&\multicolumn{2}{c|}{\textbf{WAN (1 Mb/s)}}&\multicolumn{2}{c|}{\textbf{LAN (45 Mb/s)}}&\multicolumn{2}{c|}{\textbf{WAN (1 Mb/s)}}\\\cline{2-13}
          &$T(Trans.)$&$T(Total)$&$T(Trans.)$&$T(Total)$&$T(Trans.)$&$T(Total)$&$T(Trans.)$&$T(Total)$&$T(Trans.)$&$T(Total)$&$T(Trans.)$&$T(Total)$\\
         \hline
        $\textbf{FedAvg}$&24.89&37.24&1,120.00&1,132.35&63.59&140.66&2,861.39&2,939.90&62.29&114.10&2,802.99&2,859.66\\
         $\textbf{FedProx}$&26.41&42.07&1,188.58&1,204.23&88.89&184.82&4,000.10&4,096.03&93.43&169.22&4,204.49&4,283.77\\
         ${\textbf{FedDC}}$&23.08&45.71&1,071.43&1,094.78&52.07&173.97&2,343.13&2,465.59&55.15&193.14&2,481.82&2,619.80\\
         ${\textbf{MOON}}$&22.41&40.98&1,008.57&1,028.45&48.83&148.19&2,197.14&2,297.98&64.24&177.92&2,890.58&3,004.27\\
         $\textbf{GCFL+}$&45.21&78.11&2,034.29&2,067.19&32.44&87.48&1,459.89&1,517.65&53.20&130.98&2,394.22&2,477.14\\
         $\textbf{FedStar}$&49.84&69.94&2,242.87&2,262.97&84.90&228.66&3,820.39&3,965.47&141.08&313.29&6,348.60&6,521.81\\\hline
         $\textbf{\system}$&1.24&20.82&55.61&75.20&14.56&58.04&655.19&699.87&47.80&88.41&2,050.92&2,091.09\\
         $\textbf{\advanced}$&\bf 0.65&\bf 16.02&\bf 29.23&\bf 44.61&\bf 6.16&\bf 56.79&\bf 277.28&\bf 330.91&\bf 17.00&\bf 79.25&\bf 765.15&\bf 830.39\\
         \hline
    \end{tabular}}
    \caption{{Comparison of the server-client data transition and total online training time (seconds).}}
    \label{tab:cost1}
   \vspace{-0.2in}
\end{table*}

{\bf Across-domain setting.}  
In Table \ref{tab:performance mix auc} ("Across-domain setting" column), we report global AUCs in a mixed across-domain setting. Notably, both \system\ and \advanced\ remain consistently effective under across-domain setting, with AUCs comparable to single- and across-dataset settings.
Moreover, both \system\ and \advanced\ consistently outperform all the baselines, showcasing improvements up to 23.1\% on IMDB-M. These results demonstrate the efficacy of our models in handling non-IID data. 
Furthermore, similar to the across-dataset setting, the AUC improves for certain datasets over the single-dataset setting. For example, the AUC MUTAG, ENZYMES, IMDB-M improves by 1.1\%, 15.9\%, and 3.9\%, respectively, compared to the single-dataset setting. Although the performance of IMDB-B drops slightly, the overall average AUC across all four datasets improves by 3.1\%. These results indicate that across-domain collaboration is able to enhance each other's performance. This provides an interesting point for further study.

We omit the results of graph classification accuracy under three different settings due to space limits. The observations are similar to those of the AUC performance.
\begin{figure}[t!]
    \centering
          \begin{subfigure}[b]{.36\textwidth}
         \centering
\includegraphics[width=0.75\textwidth]{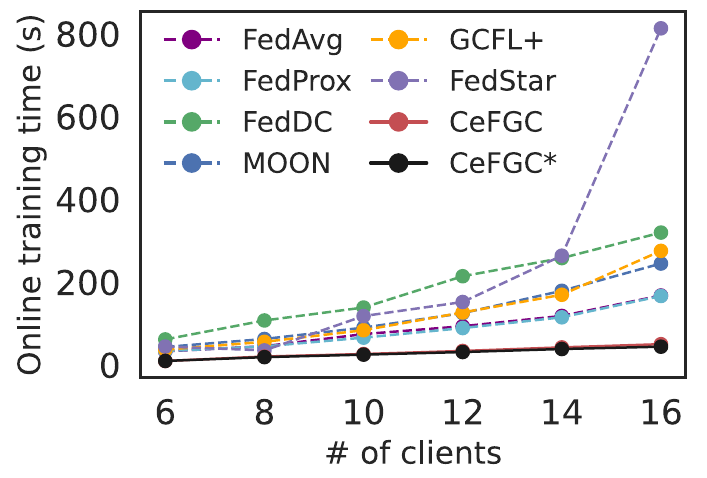}
    \end{subfigure}
    \vspace{-0.15in}
\caption{\label{fig:expand} {Comparison of 
total online running time under clients' expansion (seconds) (
MUTAG dataset, LAN-45 Mb/s).} }
 \vspace{-0.15in}
\end{figure}

{\bf Convergence of local GNN models.} As local GNNs are trained over both synthetic and local graphs which may have diverse distributions, we assess their convergence by plotting learning curves. 
We show the learning curves of local GNNs for the three settings in Appendix \ref{appx:loss curve}. The convergence condition is defined as no loss decrease within 30 continuous epochs. 
Overall, local GNNs are well-converged within 500 epochs across all settings. 
Furthermore, we evaluate local loss values on each client’s testing data with local GNN, as well as global model loss on global testing data.  
We observe that our models consistently outperform the baselines across all settings, in both average local and global loss. Due to space limits, we omit these results.
This advocates that the local models in our framework are well-converged and benefit from the generated data.

\begin{figure*}[t!]
    \centering
    \begin{tabular}{ccc}
\begin{subfigure}[b]{.3\textwidth}
      \centering
     \includegraphics[width=\textwidth]{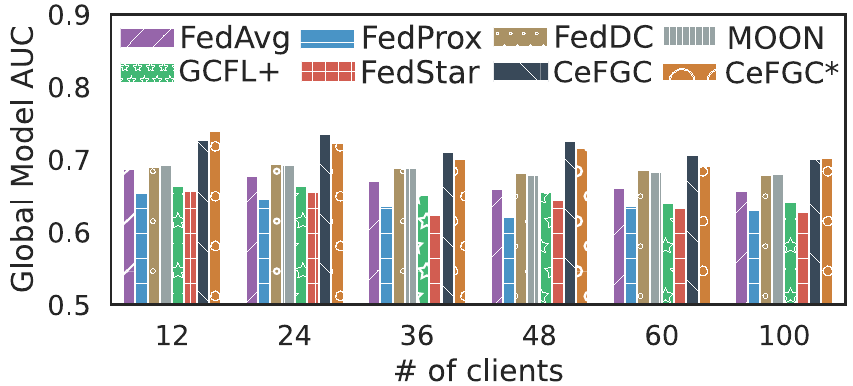}
     \vspace{-0.2in}
    \caption{\label{fig:vary-clients-all} Impact of client number.}
    \end{subfigure}
      &\begin{subfigure}[b]{.36\textwidth}
         \centering
\includegraphics[width=\textwidth]{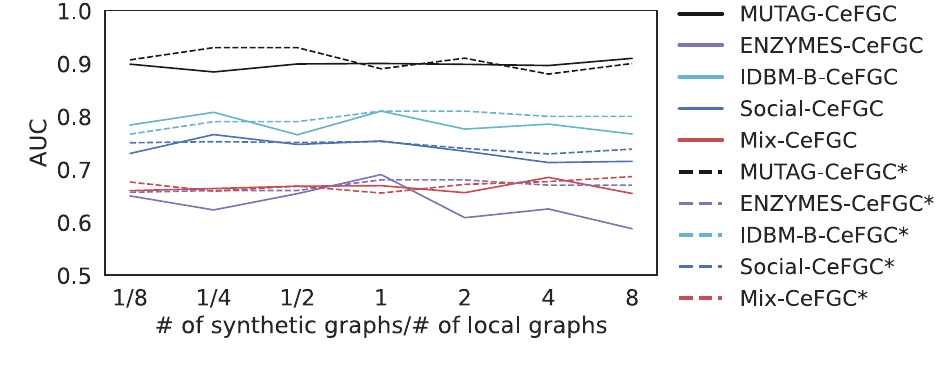}
     \vspace{-0.2in}
\caption{\label{fig:vary-synthetic-graphs-auc}Impact of synthetic graph ratio. }   
    \end{subfigure}
    &
\begin{subfigure}[b]{.3\textwidth}
      \centering
     \includegraphics[width=\textwidth]{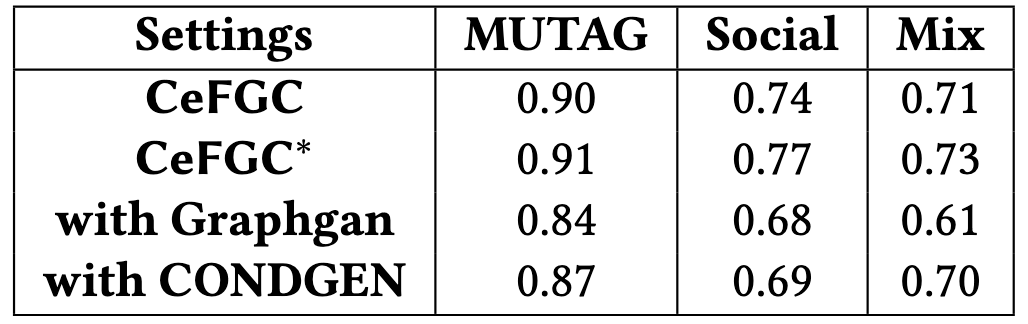}
     \vspace{-0.2in}
    \caption{\label{fig:gan} Impact of generative model.}
    \end{subfigure}
\end{tabular}
    \vspace{-0.2in}
\caption{\label{fig:impact} Impact of client numbers, synthetic graph ratios, and incorporation of GANs on global model AUC. "Social" denotes the across-dataset setting involving the IMDB-B and IMDB-M datasets, while 'Mix' refers to the across-domain setting. Figure (a) corresponds to the across-domain setting.}
\end{figure*}

\subsection{Efficiency Evaluation}
In this section, we evaluate the efficiency of all competitors in terms of \emph{communication rounds}, \emph{communication volume}, and \emph{online training time}.

{\bf Communication rounds.} 
We report the total number of server-client communications rounds for our models and the six baselines under three settings Table \ref{tab:rounds} ("Communication rounds" column). For the baselines, we count the number of communication rounds until the global model converges, defined as the point when the global loss no longer decreases over 30 consecutive training epochs. 
We observe that our models significantly reduce the communication rounds from $10^2$ to just 3, which demonstrates the superior communication efficiency of our models.

{\bf Communication volume.} 
Regarding the communication volume, recall that we provide a theoretical analysis of the data transferred per round as well as the total communication volume for both our models and the baselines in Table \ref{tab:volume}. We then empirically measure the communication costs across all three experimental settings. The results are presented in Table \ref{tab:rounds} ("Communication volume" column). We observe that our models significantly reduce the volume of data by $3$ to $10^2\times$ compared to the baselines. This reduction primarily stems from the fact that our models require only three communication rounds.

{\bf Online training time.} 
{
First, we measure the overall online training time of our model, including synthetic graphs generation, local GNN training, server-side aggregation, and client-server data transfer. We exclude the local GGDM training, as it is a one-time, offline process and the trained models is reusable across multiple FL tasks.\footnote{The training times of EDP-GNN-based diffusion models on MUTAG, ENZYMES, IMDB-B, IMDB-M, and PROTEINS are 542, 1,899, 3,537, 4,049, and 3,819 seconds, respectively. It is important to note that our framework is compatible with any GGDMs capable of capturing the original data distribution; newer models such as GDSS or Digress could further accelerate local diffusion training.} Particularly, in dynamic FL settings, existing clients can reuse their GGDMs when new clients join the collaborative learning process.
The online time costs of baselines include three main parts: local GNN training, server-side aggregation, and server-client communication cost.}
We evaluate the server-client data transfer and total time costs under two network conditions: (1) local area network (LAN) with 45 Mb/s bandwidth, (2) wide area network (WAN) with 1 Mb/s. The results in Table \ref{tab:cost1} show that our model achieves up to 98.0\% reduction in running time compared to baselines. 

Second, we evaluate the online running time under a dynamic FL setting by progressively increasing the number of clients. For a given dataset, we first evenly and randomly split graphs into 16 clients. We then randomly select an initial subset of 6 clients and expand the client pool to 8, 10, 12, 14, and 16 by incrementally adding clients from the remaining pool.
Figure \ref{fig:expand} illustrates the online running time as the number of clients increases. Our model consistently runs faster than baselines in all expanded settings.

\vspace{-0.12in}
\subsection{Factor Analysis }

In this section, we investigate the impact of the number of clients, the number of synthetic graphs, the inclusion of node feature generation, and the use of Generative Adversarial Network (GAN) for graph generation on the performance of our models. 

{\bf Varying the number of clients.} 
We vary the number of clients to evaluate scalability. We consider client counts of 12, 24, 36, 48, 60, and 100 in the across-domain setting. The first five configurations correspond to mixing datasets from MUTAG, IMDB-B, IMDB-M, and PROTEINS, with each uniformly split into 3, 6, 9, 12, and 15 clients, respectively. The 100-client setting consists of 15 clients from MUTAG, 10 from ENZYMES, and 25 each from PROTEINS, IMDB-B, and IMDB-M. 
Figure \ref{fig:vary-clients-all} reports the global model AUC on global testing data. We have the following observations. First, both our models and the baselines maintain stable performance as the client count increases, with AUC varying no more than 0.03 and 0.04 for \system\ and \advanced, respectively.
Second, our models consistently outperform the baselines across all client counts. We omit the results of the single- and across-dataset due to space limits. The observations mirror those in the across-domain setting.

{\bf Varying the number of synthetic graphs.}
We vary the ratio $r$ between the number of synthetic graphs and the number of the local graphs ($r$= 1/8, 1/4, 1/2, 1, 2, 4, 8), and evaluate the global model AUC on global testing data. From Figure \ref{fig:vary-synthetic-graphs-auc} we observe that our model maintains stable performance across all settings, and there is no consistent pattern between the AUC performance and the ratio between the number of synthetic graphs and the number of local graphs. This is expected as the performance depends on the quality and diversity of the synthetic graphs. 

{{\bf Incorporation of GANs.} As GANs have been widely used in the literature for data generation, we replace the generative model in CeFGC with two state-of-the-art graph GANs, Graphgan \cite{wang2017graph} and CONDGEN \cite{yang2019conditional}, and compare graph classification results with diffusion-based models. From Table \ref{fig:gan}, we observe that models outperform the GAN-based models. 
We believe it is contributed to the following reasons: (1) GANs are known to suffer from issues such as mode collapse and training instability, making them difficult to optimize \cite{salimans2016improved}. In contrast, diffusion models offer more stable training by gradually refining the graph structure \cite{sohl2015deep}. (2) Recent studies show that diffusion models outperform GANs in generation quality, mode coverage, and training stability \cite{karras2022elucidating,dhariwal2021diffusion,hoogeboom2023simple}.}

{\bf Effects of node feature generation.}
We have assessed the performance of our models using the GGDM of EDP-GNN that exclusively generates the graph structure. Now, we extend the evaluation of our models with GGDMs that encompass both node features and graph structure. Specifically, we utilize GDSS model \cite{jo2022score}, which jointly generates node features and adjacency matrix.
Due to space limitations, we omit the global graph classification AUC results here. Our findings indicate no significant difference in the global model performance of \system\ and \advanced\ between GGDMs that produce only graph structures and those that generate both node features and graph structures, with AUC differences below 0.02. This suggests that incorporating node feature generation has a minimal impact on the performance of \system\ and \advanced.

\begin{table*}[t!]
\begin{tabular}{cc}
    \centering
    \scalebox{0.86}{\begin{tabular}{|c|c|c|c|c|}
    \hline
\textbf{Dataset}&\textbf{\system}&\textbf{\system+}&\textbf{\advanced}&\textbf{\advanced+}\\\hline
    \textbf{PROTEINS (3 clients)} &0.87&0.86&0.87&0.86\\
    \textbf{IMDB-B (3 clients)} &0.82&0.83&0.85&0.83\\ 
    \textbf{IMDB-M (3 clients)} &0.76&0.76&0.77&0.76\\
    \textbf{Social (6 clients)} &0.77&0.74&0.77&0.75\\ 
    \textbf{Mix (12 clients)} &0.73&0.71&0.74&0.73\\\hline
    \end{tabular}}
    &    
    \scalebox{0.86}{\begin{tabular}{|c|c|c|c|c|}
    \hline
\textbf{Dataset}&\textbf{\system}&\textbf{\system+}&\textbf{\advanced}&\textbf{\advanced+}\\\hline
    \textbf{PROTEINS (15 clients)} &0.81&0.81&0.85&0.83\\
    \textbf{IMDB-B (15 clients)} &0.82&0.83&0.85&0.84\\ 
    \textbf{IMDB-M (15 clients)} &0.77&0.77&0.76&0.77\\
    \textbf{Social graphs (30 clients)} &0.77&0.77&0.76&0.76\\ 
    \textbf{Mix (60 clients)} &0.70&0.72&0.69&0.71\\\hline
    \end{tabular}}
    \end{tabular}
    \caption{\label{tab:enhance} AUC of the enhanced models. \system+ and \advanced+ denote \system\ and \advanced\ integrated with a privacy-enhancing aggregation mechanism, respectively. "Social graphs" refers to the across-dataset setting with IMDB-B and IMDB-M. "Mix" represents the across-domain setting with Proteins, IMDB-B, and IMDB-M.}
  \vspace{-0.2in}
\end{table*}

\vspace{-0.05in}
\subsection{Extension to Other Graph Learning Tasks}
{In the literature, most graph learning applications can be categorized into three major tasks: node classification, link prediction, and graph classification. Therefore, we extend our model to node classification and link prediction. Specifically, we replace the GIN in our models and baselines with VGAE \cite{kipf2016variational} as the local GNN. We split Cora \cite{sen2008collective} dataset into three subgraphs via Girvan–Newman algorithm \cite{girvan2002community}, assigning one to each client. Our framework remains effective for link prediction, achieving an AUC of 0.81. Regarding the node classification, since current diffusion-based generators cannot produce node labels, our model does not yet support this task. This limitation highlights labeled graph generation as a promising and important future work.}

\vspace{-0.05in}
\section{Enhancing the Privacy of Our Model}
\label{sc:enhanced}
In both \system\ and \advanced, we enhance the privacy of our model in the following three ways: First, we reduce the number of server-client communication rounds to three, which mitigates the effectiveness of inference attacks \cite{wang2019beyond, hitaj2017deep, shokri2017membership, melis2019exploiting, zhu2019deep, zhao2020idlg} and poisoning attacks \cite{yin2021comprehensive, bouacida2021vulnerabilities, lyu2020threats, mothukuri2021survey} against FL models, as these attacks typically depend on analyzing gradient or embedding differences across multiple communication rounds. Second, 
we incorporate encryption techniques in our model that each client shares an encrypted diffusion model with the server, ensuring the privacy of local data. 
Third, the server randomly shuffles the diffusion models before distributing them to the clients, ensuring that the diffusion models cannot be linked to their respective IDs. 
However, some curious clients may still attempt to infer data information from the generative models distributed by the server, even though they cannot associate the diffusion models with specific client IDs. Therefore, to enhance the privacy of our model, we employ a key strategy: the server aggregates the generative models within the encryption domain before distributing them to the clients.

First, we need to ensure that computations performed on encrypted data on the server yield the same results as those applied to the original (decrypted) data while providing strong privacy guarantees. 
{To achieve this, we follow previous works \cite{roth2022nvidia, liu2021fate, ludwig2020ibm, stripelis2021secure}, allowing the clients in our model to apply Homomorphic Encryption (HE) \cite{gentry2009fully, acar2018survey} to encrypt their generative models. HE is popularly adopted in both FL research \cite{park2022privacy, zhang2020batchcrypt, xie2024efficiency, ma2022privacy, fang2021privacy} and related applications such as banking \cite{liu2021fate}, Internet of Things (IoT) systems \cite{zhang2022homomorphic, hijazi2023secure}, and medical data analysis \cite{zhou2024personalized, gkillas2024privacy, hardy2017private}. We note that client-server collusion is not considered in this study, as it falls beyond the scope of our work.}
Since HE ensures equivalent results for computations on both encrypted and decrypted data, we describe the aggregation mechanism as it occurs in the decrypted domain for clarity.
The aggregation mechanism proceeds as follows:

First, for each pair of generative model $M_i$ and $M_j$, where $i \neq j$, the server measures the similarity $sim(M_i, M_j)$ between $M_i$ and $M_j$ with cosine similarity, which can be formulated as:
\vspace{-0.05in}
 \begin{equation}
     \label{eqn:score-dis}
     \begin{split}
     sim(M_i, M_j) = \textit{cos}(s_{\theta_i}, s_{\theta_j}),
     \end{split}
 \end{equation}
where $s$ represents the conditional score model, $\theta_i$ and $\theta_j$ are the parameters of the conditional score models for $M_i$ and $M_j$, respectively. Intuitively, a higher $sim()$ value indicates that the $M_i$ and $M_j$ are more similar to each other. 

Second, for each generative model $M_i$, the server identifies a $M_j$ which is most similar to $M_i$ (i.e., with the largest $sim(M_i, M_j)$). It then averages $M_i$ and $M_j$ to update the parameters of $M_i$, resulting in an aggregated model $\hat{M}_i$ calculated as:
\vspace{-0.05in}
 \begin{equation}
     \label{eqn:score-agg}
     \small
     \begin{split}
     \hat{M}_i = \frac{1}{2}(s_{\theta_i}+s_{\theta_j}).
     \end{split}
 \end{equation}
 \vspace{-0.15in}
 
After obtaining $\hat{M}_i$, for each client, the server packs the aggregated generative models from all clients, excluding the client’s own aggregated model. The packed models are randomly shuffled and distributed back to the clients, following the same process described in Phase 2 of Section \ref{sc:model}.

Finally, we evaluate the global model AUC performance of \system\ and \advanced\ with the privacy-enhanced method, denoted as \system+ and \advanced+. The evaluation is conducted under single-dataset, across-dataset, and across-domain settings. 
The results in Table \ref{tab:enhance} show that the performance of \system+ and \advanced+ is comparable to that of \system\ and \advanced, respectively, demonstrating the effectiveness of the enhanced mechanism.

\section{Conclusion}
\label{sec:conclusion}

Federated GNNs (FGNNs) tackle decentralized graph learning but face two important challenges: i) high communication overhead due to frequent parameter exchanges; ii) non-IID local data among clients. To address these issues, we propose \system\ and \advanced, a novel type of FGNN paradigm that enables training over non-IID data with just three rounds of communication by utilizing generative models.
\system\ and \advanced\ use generative models to minimize direct client-server communication. Each client trains a generative model with or without a graph label channel, encrypts it, and shares it with the server. The server then randomly shuffles and aggregates the uploaded generative models and distributes them back to the clients. Clients generate synthetic graphs based on the generative models from the server, train local GNNs, and upload weights to the server for global aggregation. 
Extensive experiments and analysis demonstrate the effectiveness and efficiency of our models across diverse settings.

\section{Acknowledgments}
We thank the anonymous reviewers for their feedback. This work was supported by the Hong Kong RGC Grants C2003-23Y and 12201925.
\newpage
\bibliographystyle{plain}
\bibliography{bib}

@article{acar2018survey,
  title={A survey on homomorphic encryption schemes: Theory and implementation},
  author={Acar, Abbas and Aksu, Hidayet and Uluagac, A Selcuk and Conti, Mauro},
  journal={ACM Computing Surveys (Csur)},
  volume={51},
  number={4},
  pages={1--35},
  year={2018},
}

@article{hardy2017private,
  title={Private federated learning on vertically partitioned data via entity resolution and additively homomorphic encryption},
  author={Hardy, Stephen and Henecka, Wilko and Ivey-Law, Hamish and Nock, Richard and Patrini, Giorgio and Smith, Guillaume and Thorne, Brian},
  journal={arXiv preprint arXiv:1711.10677},
  year={2017}
}

@article{huang2013depth,
  title={An in-depth study of LTE: Effect of network protocol and application behavior on performance},
  author={Huang, Junxian and Qian, Feng and Guo, Yihua and Zhou, Yuanyuan and Xu, Qiang and Mao, Z Morley and Sen, Subhabrata and Spatscheck, Oliver},
  journal={ACM SIGCOMM Computer Communication Review},
  volume={43},
  number={4},
  pages={363--374},
  year={2013},
}

@article{kairouz2021advances,
  title={Advances and open problems in federated learning},
  author={Kairouz, Peter and McMahan, H Brendan and Avent, Brendan and Bellet, Aur{\'e}lien and Bennis, Mehdi and Bhagoji, Arjun Nitin and Bonawitz, Kallista and Charles, Zachary and Cormode, Graham and Cummings, Rachel and others},
  journal={Foundations and trends{\textregistered} in machine learning},
  volume={14},
  number={1--2},
  pages={1--210},
  year={2021},
}

@article{kipf2016variational,
  title={Variational graph auto-encoders},
  author={Kipf, Thomas N and Welling, Max},
  journal={arXiv preprint arXiv:1611.07308},
  year={2016}
}

@article{sen2008collective,
  title={Collective classification in network data},
  author={Sen, Prithviraj and Namata, Galileo and Bilgic, Mustafa and Getoor, Lise and Galligher, Brian and Eliassi-Rad, Tina},
  journal={AI magazine},
  volume={29},
  number={3},
  pages={93--93},
  year={2008}
}

@article{girvan2002community,
  title={Community structure in social and biological networks},
  author={Girvan, Michelle and Newman, Mark EJ},
  journal={Proceedings of the national academy of sciences},
  volume={99},
  number={12},
  pages={7821--7826},
  year={2002},
}

@inproceedings{gkillas2024privacy,
  title={Privacy-preserving federated deep-equilibrium learning for medical image classification},
  author={Gkillas, Alexandros and Ampeliotis, Dimitris and Berberidis, Kostas},
  booktitle={IEEE International Symposium on Biomedical Imaging},
  pages={1--4},
  year={2024},
}

@article{zhou2024personalized,
  title={Personalized and privacy-preserving federated heterogeneous medical image analysis with PPPML-HMI},
  author={Zhou, Juexiao and Zhou, Longxi and Wang, Di and Xu, Xiaopeng and Li, Haoyang and Chu, Yuetan and Han, Wenkai and Gao, Xin},
  journal={Computers in Biology and Medicine},
  volume={169},
  pages={107861},
  year={2024},
  publisher={Elsevier}
}

@inproceedings{stripelis2021secure,
  title={Secure neuroimaging analysis using federated learning with homomorphic encryption},
  author={Stripelis, Dimitris and Saleem, Hamza and Ghai, Tanmay and Dhinagar, Nikhil and Gupta, Umang and Anastasiou, Chrysovalantis and Ver Steeg, Greg and Ravi, Srivatsan and Naveed, Muhammad and Thompson, Paul M and others},
  booktitle={International Symposium on medical information processing and analysis},
  volume={12088},
  pages={351--359},
  year={2021},
  organization={SPIE}
}

@article{ludwig2020ibm,
  title={Ibm federated learning: an enterprise framework white paper v0. 1},
  author={Ludwig, Heiko and Baracaldo, Nathalie and Thomas, Gegi and Zhou, Yi and Anwar, Ali and Rajamoni, Shashank and Ong, Yuya and Radhakrishnan, Jayaram and Verma, Ashish and Sinn, Mathieu and others},
  journal={arXiv preprint arXiv:2007.10987},
  year={2020}
}

@article{roth2022nvidia,
  title={Nvidia flare: Federated learning from simulation to real-world},
  author={Roth, Holger R and Cheng, Yan and Wen, Yuhong and Yang, Isaac and Xu, Ziyue and Hsieh, Yuan-Ting and Kersten and others},
  journal={arXiv preprint arXiv:2210.13291},
  year={2022}
}

@article{hijazi2023secure,
  title={Secure federated learning with fully homomorphic encryption for iot communications},
  author={Hijazi, Neveen Mohammad and Aloqaily, Moayad and Guizani, Mohsen and Ouni, Bassem and Karray, Fakhri},
  journal={IEEE Internet of Things Journal},
  volume={11},
  number={3},
  pages={4289--4300},
  year={2023},
}

@article{zhang2022homomorphic,
  title={Homomorphic encryption-based privacy-preserving federated learning in IoT-enabled healthcare system},
  author={Zhang, Li and Xu, Jianbo and Vijayakumar, Pandi and Sharma, Pradip Kumar and Ghosh, Uttam},
  journal={IEEE transactions on network science and engineering},
  volume={10},
  number={5},
  pages={2864--2880},
  year={2022},
}

@article{ma2022privacy,
  title={Privacy-preserving federated learning based on multi-key homomorphic encryption},
  author={Ma, Jing and Naas, Si-Ahmed and Sigg, Stephan and Lyu, Xixiang},
  journal={International Journal of Intelligent Systems},
  volume={37},
  number={9},
  pages={5880--5901},
  year={2022},
}

@article{fang2021privacy,
  title={Privacy preserving machine learning with homomorphic encryption and federated learning},
  author={Fang, Haokun and Qian, Quan},
  journal={Future Internet},
  volume={13},
  number={4},
  pages={94},
  year={2021},
}

@article{liu2021fate,
  title={Fate: An industrial grade platform for collaborative learning with data protection},
  author={Liu, Yang and Fan, Tao and Chen, Tianjian and Xu, Qian and Yang, Qiang},
  journal={Journal of Machine Learning Research},
  volume={22},
  number={226},
  pages={1--6},
  year={2021}
}

@article{salimans2016improved,
  title={Improved techniques for training gans},
  author={Salimans, Tim and Goodfellow, Ian and Zaremba, Wojciech and Cheung, Vicki and Radford, Alec and Chen, Xi},
  journal={Advances in neural information processing systems},
  volume={29},
  year={2016}
}

@inproceedings{sohl2015deep,
  title={Deep unsupervised learning using nonequilibrium thermodynamics},
  author={Sohl-Dickstein, Jascha and Weiss, Eric and Maheswaranathan, Niru and Ganguli, Surya},
  booktitle={International conference on machine learning},
  pages={2256--2265},
  year={2015},
  organization={pmlr}
}

@article{dhariwal2021diffusion,
  title={Diffusion models beat gans on image synthesis},
  author={Dhariwal, Prafulla and Nichol, Alexander},
  journal={Advances in neural information processing systems},
  volume={34},
  pages={8780--8794},
  year={2021}
}

@article{karras2022elucidating,
  title={Elucidating the design space of diffusion-based generative models},
  author={Karras, Tero and Aittala, Miika and Aila, Timo and Laine, Samuli},
  journal={Advances in neural information processing systems},
  volume={35},
  pages={26565--26577},
  year={2022}
}

@inproceedings{hoogeboom2023simple,
  title={simple diffusion: End-to-end diffusion for high resolution images},
  author={Hoogeboom, Emiel and Heek, Jonathan and Salimans, Tim},
  booktitle={International Conference on Machine Learning},
  pages={13213--13232},
  year={2023},
  organization={PMLR}
}

@article{wang2017graph,
  title={Graph representation learning with generative adversarial nets},
  author={Wang, H and Wang, J and Wang, J and others},
  journal={AAAI, Graphgan},
  year={2017}
}

@article{yang2019conditional,
  title={Conditional structure generation through graph variational generative adversarial nets},
  author={Yang, Carl and Zhuang, Peiye and Shi, Wenhan and Luu, Alan and Li, Pan},
  journal={Advances in neural information processing systems},
  volume={32},
  year={2019}
}

@inproceedings{li2021model,
  title={Model-contrastive federated learning},
  author={Li, Qinbin and He, Bingsheng and Song, Dawn},
  booktitle={Proceedings of the IEEE/CVF conference on computer vision and pattern recognition},
  pages={10713--10722},
  year={2021}
}

@inproceedings{gao2022feddc,
  title={Feddc: Federated learning with non-iid data via local drift decoupling and correction},
  author={Gao, Liang and Fu, Huazhu and Li, Li and Chen, Yingwen and Xu, Ming and Xu, Cheng-Zhong},
  booktitle={Proceedings of the IEEE/CVF conference on computer vision and pattern recognition},
  pages={10112--10121},
  year={2022}
}

@article{park2022privacy,
  title={Privacy-preserving federated learning using homomorphic encryption},
  author={Park, Jaehyoung and Lim, Hyuk},
  journal={Applied Sciences},
  volume={12},
  number={2},
  pages={734},
  year={2022},
}

@inproceedings{zhang2020batchcrypt,
  title={$\{$BatchCrypt$\}$: Efficient homomorphic encryption for $\{$Cross-Silo$\}$ federated learning},
  author={Zhang, Chengliang and Li, Suyi and Xia, Junzhe and Wang, Wei and Yan, Feng and Liu, Yang},
  booktitle={USENIX annual technical conference},
  pages={493--506},
  year={2020}
}

@article{xie2024efficiency,
  title={Efficiency optimization techniques in privacy-preserving federated learning with homomorphic encryption: A brief survey},
  author={Xie, Qipeng and Jiang, Siyang and Jiang, Linshan and Huang, Yongzhi and Zhao, Zhihe and Khan, Salabat and Dai, Wangchen and Liu, Zhe and Wu, Kaishun},
  journal={IEEE Internet of Things Journal},
  volume={11},
  number={14},
  pages={24569--24580},
  year={2024},
}

@inproceedings{gentry2009fully,
  title={Fully homomorphic encryption using ideal lattices},
  author={Gentry, Craig},
  booktitle={Proceedings of the ACM symposium on Theory of computing},
  pages={169--178},
  year={2009}
}

@article{wu2020comprehensive,
  title={A comprehensive survey on graph neural networks},
  author={Wu, Zonghan and Pan, Shirui and Chen, Fengwen and Long, Guodong and Zhang, Chengqi and Philip, S Yu},
  journal={IEEE transactions on neural networks and learning systems},
  volume={32},
  number={1},
  pages={4--24},
  year={2020},
  publisher={IEEE}
}

@article{kudo2004application,
  title={An application of boosting to graph classification},
  author={Kudo, Taku and Maeda, Eisaku and Matsumoto, Yuji},
  journal={Advances in neural information processing systems},
  volume={17},
  year={2004}
}

@article{wu2021fast,
  title={Fast-convergent federated learning with adaptive weighting},
  author={Wu, Hongda and Wang, Ping},
  journal={IEEE Transactions on Cognitive Communications and Networking},
  volume={7},
  number={4},
  pages={1078--1088},
  year={2021},
}

@article{cho2020client,
  title={Client selection in federated learning: Convergence analysis and power-of-choice selection strategies},
  author={Cho, Yae Jee and Wang, Jianyu and Joshi, Gauri},
  journal={arXiv preprint arXiv:2010.01243},
  year={2020}
}

@article{nguyen2020fast,
  title={Fast-convergent federated learning},
  author={Nguyen, Hung T and Sehwag, Vikash and Hosseinalipour, Seyyedali and Brinton, Christopher G and Chiang, Mung and Poor, H Vincent},
  journal={IEEE Journal on Selected Areas in Communications},
  volume={39},
  number={1},
  pages={201--218},
  year={2020},
}

@inproceedings{kamp2019efficient,
  title={Efficient decentralized deep learning by dynamic model averaging},
  author={Kamp, Michael and Adilova, Linara and Sicking, Joachim and H{\"u}ger, Fabian and Schlicht, Peter and Wirtz, Tim and Wrobel, Stefan},
  booktitle={Joint European Conference on Machine Learning and
Knowledge Discovery in Databases},
  pages={393--409},
  year={2019},
  organization={Springer}
}

@article{xu2020client,
  title={Client selection and bandwidth allocation in wireless federated learning networks: A long-term perspective},
  author={Xu, Jie and Wang, Heqiang},
  journal={IEEE Transactions on Wireless Communications},
  volume={20},
  number={2},
  pages={1188--1200},
  year={2020},
}

@article{abdulrahman2020fedmccs,
  title={FedMCCS: Multicriteria client selection model for optimal IoT federated learning},
  author={AbdulRahman, Sawsan and Tout, Hanine and Mourad, Azzam and Talhi, Chamseddine},
  journal={IEEE Internet of Things Journal},
  volume={8},
  number={6},
  pages={4723--4735},
  year={2020},
}

@inproceedings{reisizadeh2020fedpaq,
  title={Fedpaq: A communication-efficient federated learning method with periodic averaging and quantization},
  author={Reisizadeh, Amirhossein and Mokhtari, Aryan and Hassani, Hamed and Jadbabaie, Ali and Pedarsani, Ramtin},
  booktitle={International Conference on Artificial Intelligence and Statistics},
  pages={2021--2031},
  year={2020},
}

@inproceedings{rothchild2020fetchsgd,
  title={Fetchsgd: Communication-efficient federated learning with sketching},
  author={Rothchild, Daniel and Panda, Ashwinee and Ullah, Enayat and Ivkin, Nikita and Stoica, Ion and Braverman, Vladimir and Gonzalez, Joseph and Arora, Raman},
  booktitle={International Conference on Machine Learning},
  pages={8253--8265},
  year={2020},
}

@article{chen2021communication,
  title={Communication-efficient federated learning},
  author={Chen, Mingzhe and Shlezinger, Nir and Poor, H Vincent and Eldar, Yonina C and Cui, Shuguang},
  journal={Proceedings of the National Academy of Sciences},
  volume={118},
  number={17},
  pages={e2024789118},
  year={2021},
}

@inproceedings{ivanov2018anonymous,
  title={Anonymous walk embeddings},
  author={Ivanov, Sergey and Burnaev, Evgeny},
  booktitle={International conference on machine learning},
  pages={2186--2195},
  year={2018},
}

@article{kingma2014adam,
  title={Adam: A method for stochastic optimization},
  author={Kingma, Diederik P and Ba, Jimmy},
  journal={arXiv preprint arXiv:1412.6980},
  year={2014}
}

@article{ruder2016overview,
  title={An overview of gradient descent optimization algorithms},
  author={Ruder, Sebastian},
  journal={arXiv preprint arXiv:1609.04747},
  year={2016}
}

@article{liu2022federated,
  title={Federated graph neural networks: Overview, techniques and challenges},
  author={Liu, Rui and Xing, Pengwei and Deng, Zichao and Li, Anran and Guan, Cuntai and Yu, Han},
  journal={arXiv preprint arXiv:2202.07256},
  year={2022}
}

@article{ying2018hierarchical,
  title={Hierarchical graph representation learning with differentiable pooling},
  author={Ying, Zhitao and You, Jiaxuan and Morris, Christopher and Ren, Xiang and Hamilton, Will and Leskovec, Jure},
  journal={Advances in neural information processing systems},
  volume={31},
  year={2018}
}

@inproceedings{zhang2018end,
  title={An end-to-end deep learning architecture for graph classification},
  author={Zhang, Muhan and Cui, Zhicheng and Neumann, Marion and Chen, Yixin},
  booktitle={Proceedings of the AAAI conference on artificial intelligence},
  volume={32},
  number={1},
  year={2018}
}

@inproceedings{guo2023globally,
  title={Globally consistent federated graph autoencoder for non-IID graphs},
  author={Guo, Kun and Fang, Yutong and Huang, Qingqing and Liang, Yuting and Zhang, Ziyao and He, Wenyu and Yang, Liu and Chen, Kai and Liu, Ximeng and Guo, Wenzhong},
  booktitle={Proceedings of the International Joint Conference on Artificial Intelligence},
  pages={3768--3776},
  year={2023}
}

@article{sattler2019robust,
  title={Robust and communication-efficient federated learning from non-iid data},
  author={Sattler, Felix and Wiedemann, Simon and M{\"u}ller, Klaus-Robert and Samek, Wojciech},
  journal={IEEE transactions on neural networks and learning systems},
  volume={31},
  number={9},
  pages={3400--3413},
  year={2019},
}

@article{konevcny2016federated,
  title={Federated learning: Strategies for improving communication efficiency},
  author={Kone{\v{c}}n{\`y}, Jakub and McMahan, H Brendan and Yu, Felix X and Richt{\'a}rik, Peter and Suresh, Ananda Theertha and Bacon, Dave},
  journal={arXiv preprint arXiv:1610.05492},
  year={2016}
}

@article{zhao2020idlg,
  title={idlg: Improved deep leakage from gradients},
  author={Zhao, Bo and Mopuri, Konda Reddy and Bilen, Hakan},
  journal={arXiv preprint arXiv:2001.02610},
  year={2020}
}

@article{zhu2019deep,
  title={Deep leakage from gradients},
  author={Zhu, Ligeng and Liu, Zhijian and Han, Song},
  journal={Advances in neural information processing systems},
  volume={32},
  year={2019}
}

@article{yin2021comprehensive,
  title={A comprehensive survey of privacy-preserving federated learning: A taxonomy, review, and future directions},
  author={Yin, Xuefei and Zhu, Yanming and Hu, Jiankun},
  journal={ACM Computing Surveys},
  volume={54},
  number={6},
  pages={1--36},
  year={2021},
}

@inproceedings{shokri2017membership,
  title={Membership inference attacks against machine learning models},
  author={Shokri, Reza and Stronati, Marco and Song, Congzheng and Shmatikov, Vitaly},
  booktitle={IEEE symposium on security and privacy},
  pages={3--18},
  year={2017},
}

@inproceedings{melis2019exploiting,
  title={Exploiting unintended feature leakage in collaborative learning},
  author={Melis, Luca and Song, Congzheng and De Cristofaro, Emiliano and Shmatikov, Vitaly},
  booktitle={IEEE symposium on security and privacy},
  pages={691--706},
  year={2019},
}

@inproceedings{wang2019beyond,
  title={Beyond inferring class representatives: User-level privacy leakage from federated learning},
  author={Wang, Zhibo and Song, Mengkai and Zhang, Zhifei and Song, Yang and Wang, Qian and Qi, Hairong},
  booktitle={IEEE INFOCOM conference on computer communications},
  pages={2512--2520},
  year={2019},
}

@inproceedings{hitaj2017deep,
  title={Deep models under the GAN: information leakage from collaborative deep learning},
  author={Hitaj, Briland and Ateniese, Giuseppe and Perez-Cruz, Fernando},
  booktitle={Proceedings of the ACM SIGSAC conference on computer and communications security},
  pages={603--618},
  year={2017}
}

@article{bouacida2021vulnerabilities,
  title={Vulnerabilities in federated learning},
  author={Bouacida, Nader and Mohapatra, Prasant},
  journal={IEEE Access},
  volume={9},
  pages={63229--63249},
  year={2021},
}

@article{mothukuri2021survey,
  title={A survey on security and privacy of federated learning},
  author={Mothukuri, Viraaji and Parizi, Reza M and Pouriyeh, Seyedamin and Huang, Yan and Dehghantanha, Ali and Srivastava, Gautam},
  journal={Future Generation Computer Systems},
  volume={115},
  pages={619--640},
  year={2021},
}

@article{lyu2020threats,
  title={Threats to federated learning: A survey},
  author={Lyu, Lingjuan and Yu, Han and Yang, Qiang},
  journal={arXiv preprint arXiv:2003.02133},
  year={2020}
}

@article{zhou2020distilled,
  title={Distilled one-shot federated learning},
  author={Zhou, Yanlin and Pu, George and Ma, Xiyao and Li, Xiaolin and Wu, Dapeng},
  journal={arXiv preprint arXiv:2009.07999},
  year={2020}
}

@article{guha2019one,
  title={One-shot federated learning},
  author={Guha, Neel and Talwalkar, Ameet and Smith, Virginia},
  journal={arXiv preprint arXiv:1902.11175},
  year={2019}
}

@article{yang2023diffusion,
  title={Diffusion models: A comprehensive survey of methods and applications},
  author={Yang, Ling and Zhang, Zhilong and Song, Yang and Hong, Shenda and Xu, Runsheng and Zhao, Yue and Zhang, Wentao and Cui, Bin and Yang, Ming-Hsuan},
  journal={ACM Computing Surveys},
  volume={56},
  number={4},
  pages={1--39},
  year={2023},
}

@article{cao2022survey,
  title={A survey on generative diffusion model},
  author={Cao, Hanqun and Tan, Cheng and Gao, Zhangyang and Chen, Guangyong and Heng, Pheng-Ann and Li, Stan Z},
  journal={arXiv preprint arXiv:2209.02646},
  year={2022}
}

@inproceedings{zhang2021graph,
  title={Graph embedding for recommendation against attribute inference attacks},
  author={Zhang, Shijie and Yin, Hongzhi and Chen, Tong and Huang, Zi and Cui, Lizhen and Zhang, Xiangliang},
  booktitle={Proceedings of the Web Conference 2021},
  pages={3002--3014},
  year={2021}
}

@article{min2021stgsn,
  title={STGSN—A Spatial--Temporal Graph Neural Network framework for time-evolving social networks},
  author={Min, Shengjie and Gao, Zhan and Peng, Jing and Wang, Liang and Qin, Ke and Fang, Bo},
  journal={Knowledge-Based Systems},
  volume={214},
  pages={106746},
  year={2021},
}

@article{jin2021application,
  title={Application of deep learning methods in biological networks},
  author={Jin, Shuting and Zeng, Xiangxiang and Xia, Feng and Huang, Wei and Liu, Xiangrong},
  journal={Briefings in bioinformatics},
  volume={22},
  number={2},
  pages={1902--1917},
  year={2021},
}

@article{gao2023survey,
  title={A survey of graph neural networks for recommender systems: Challenges, methods, and directions},
  author={Gao, Chen and Zheng, Yu and Li, Nian and Li, Yinfeng and Qin, Yingrong and Piao, Jinghua and Quan, Yuhan and Chang, Jianxin and Jin, Depeng and He, Xiangnan and others},
  journal={ACM Transactions on Recommender Systems},
  volume={1},
  number={1},
  pages={1--51},
  year={2023},
}

@inproceedings{gao2022graph,
  title={Graph neural networks for recommender system},
  author={Gao, Chen and Wang, Xiang and He, Xiangnan and Li, Yong},
  booktitle={Proceedings of the Fifteenth ACM International Conference on Web Search and Data Mining},
  pages={1623--1625},
  year={2022}
}

@inproceedings{fan2019graph,
  title={Graph neural networks for social recommendation},
  author={Fan, Wenqi and Ma, Yao and Li, Qing and He, Yuan and Zhao, Eric and Tang, Jiliang and Yin, Dawei},
  booktitle={The world wide web conference},
  pages={417--426},
  year={2019}
}

@article{dwivedi2021graph,
  title={Graph neural networks with learnable structural and positional representations},
  author={Dwivedi, Vijay Prakash and Luu, Anh Tuan and Laurent, Thomas and Bengio, Yoshua and Bresson, Xavier},
  journal={arXiv preprint arXiv:2110.07875},
  year={2021}
}

@inproceedings{tan2023federated,
  title={Federated learning on non-iid graphs via structural knowledge sharing},
  author={Tan, Yue and Liu, Yixin and Long, Guodong and Jiang, Jing and Lu, Qinghua and Zhang, Chengqi},
  booktitle={Proceedings of the AAAI conference on artificial intelligence},
  volume={37},
  number={8},
  pages={9953--9961},
  year={2023}
}

@article{wu2021fedgnn,
  title={Fedgnn: Federated graph neural network for privacy-preserving recommendation},
  author={Wu, Chuhan and Wu, Fangzhao and Cao, Yang and Huang, Yongfeng and Xie, Xing},
  journal={arXiv preprint arXiv:2102.04925},
  year={2021}
}

@article{shahid2021communication,
  title={Communication efficiency in federated learning: Achievements and challenges},
  author={Shahid, Osama and Pouriyeh, Seyedamin and Parizi, Reza M and Sheng, Quan Z and Srivastava, Gautam and Zhao, Liang},
  journal={arXiv preprint arXiv:2107.10996},
  year={2021}
}

@article{xie2021federated,
  title={Federated graph classification over non-iid graphs},
  author={Xie, Han and Ma, Jing and Xiong, Li and Yang, Carl},
  journal={Advances in Neural Information Processing Systems},
  volume={34},
  pages={18839--18852},
  year={2021}
}

@article{li2020federated,
  title={Federated optimization in heterogeneous networks},
  author={Li, Tian and Sahu, Anit Kumar and Zaheer, Manzil and Sanjabi, Maziar and Talwalkar, Ameet and Smith, Virginia},
  journal={Proceedings of Machine learning and systems},
  volume={2},
  pages={429--450},
  year={2020}
}

@article{song2019generative,
  title={Generative modeling by estimating gradients of the data distribution},
  author={Song, Yang and Ermon, Stefano},
  journal={Advances in neural information processing systems},
  volume={32},
  year={2019}
}

@article{morris2020tudataset,
  title={Tudataset: A collection of benchmark datasets for learning with graphs},
  author={Morris, Christopher and Kriege, Nils M and Bause, Franka and Kersting, Kristian and Mutzel, Petra and Neumann, Marion},
  journal={arXiv preprint arXiv:2007.08663},
  year={2020}
}

@article{xu2018powerful,
  title={How powerful are graph neural networks?},
  author={Xu, Keyulu and Hu, Weihua and Leskovec, Jure and Jegelka, Stefanie},
  journal={arXiv preprint arXiv:1810.00826},
  year={2018}
}

@article{zhou2020graph,
  title={Graph neural networks: A review of methods and applications},
  author={Zhou, Jie and Cui, Ganqu and Hu, Shengding and Zhang, Zhengyan and Yang, Cheng and Liu, Zhiyuan and Wang, Lifeng and Li, Changcheng and Sun, Maosong},
  journal={AI open},
  volume={1},
  pages={57--81},
  year={2020},
}

@article{haefeli2022diffusion,
  title={Diffusion models for graphs benefit from discrete state spaces},
  author={Haefeli, Kilian Konstantin and Martinkus, Karolis and Perraudin, Nathana{\"e}l and Wattenhofer, Roger},
  journal={arXiv preprint arXiv:2210.01549},
  year={2022}
}

@article{vignac2022digress,
  title={Digress: Discrete denoising diffusion for graph generation},
  author={Vignac, Clement and Krawczuk, Igor and Siraudin, Antoine and Wang, Bohan and Cevher, Volkan and Frossard, Pascal},
  journal={arXiv preprint arXiv:2209.14734},
  year={2022}
}

@article{luo2022fast,
  title={Fast graph generative model via spectral diffusion},
  author={Luo, Tianze and Mo, Zhanfeng and Pan, Sinno Jialin},
  journal={arXiv preprint arXiv:2211.08892},
  year={2022}
}

@inproceedings{jo2022score,
  title={Score-based generative modeling of graphs via the system of stochastic differential equations},
  author={Jo, Jaehyeong and Lee, Seul and Hwang, Sung Ju},
  booktitle={International Conference on Machine Learning},
  pages={10362--10383},
  year={2022},
}

@inproceedings{huang2022graphgdp,
  title={Graphgdp: Generative diffusion processes for permutation invariant graph generation},
  author={Huang, Han and Sun, Leilei and Du, Bowen and Fu, Yanjie and Lv, Weifeng},
  booktitle={IEEE International Conference on Data Mining (ICDM)},
  pages={201--210},
  year={2022},
}

@article{chen2022nvdiff,
  title={Nvdiff: Graph generation through the diffusion of node vectors},
  author={Chen, Xiaohui and Li, Yukun and Zhang, Aonan and Liu, Li-ping},
  journal={arXiv preprint arXiv:2211.10794},
  year={2022}
}

@inproceedings{niu2020permutation,
  title={Permutation invariant graph generation via score-based generative modeling},
  author={Niu, Chenhao and Song, Yang and Song, Jiaming and Zhao, Shengjia and Grover, Aditya and Ermon, Stefano},
  booktitle={International Conference on Artificial Intelligence and Statistics},
  pages={4474--4484},
  year={2020},
}

@article{zhang2023survey,
  title={A survey on graph diffusion models: Generative ai in science for molecule, protein and material},
  author={Zhang, Mengchun and Qamar, Maryam and Kang, Taegoo and Jung, Yuna and Zhang, Chenshuang and Bae, Sung-Ho and Zhang, Chaoning},
  journal={arXiv preprint arXiv:2304.01565},
  year={2023}
}

@article{austin2021structured,
  title={Structured denoising diffusion models in discrete state-spaces},
  author={Austin, Jacob and Johnson, Daniel D and Ho, Jonathan and Tarlow, Daniel and Van Den Berg, Rianne},
  journal={Advances in Neural Information Processing Systems},
  volume={34},
  pages={17981--17993},
  year={2021}
}

@article{ho2020denoising,
  title={Denoising diffusion probabilistic models},
  author={Ho, Jonathan and Jain, Ajay and Abbeel, Pieter},
  journal={Advances in neural information processing systems},
  volume={33},
  pages={6840--6851},
  year={2020}
}

@article{fu2022federated,
  title={Federated graph machine learning: A survey of concepts, techniques, and applications},
  author={Fu, Xingbo and Zhang, Binchi and Dong, Yushun and Chen, Chen and Li, Jundong},
  journal={ACM SIGKDD Explorations Newsletter},
  volume={24},
  number={2},
  pages={32--47},
  year={2022},
}

@article{he2021fedgraphnn,
  title={Fedgraphnn: A federated learning system and benchmark for graph neural networks},
  author={He, Chaoyang and Balasubramanian, Keshav and Ceyani, Emir and Yang, Carl and Xie, Han and Sun, Lichao and He, Lifang and Yang, Liangwei and Yu, Philip S and Rong, Yu and others},
  journal={arXiv preprint arXiv:2104.07145},
  year={2021}
}

@article{mcmahan2017communication,
  title={Communication-efficient learning of deep networks from decentralized data},
  author={McMahan, Brendan and Moore, Eider and Ramage, Daniel and Hampson, Seth and y Arcas, Blaise Aguera},
  booktitle={Artificial intelligence and statistics},
  pages={1273--1282},
  year={2017}
}

@article{he2020fedml,
  title={Fedml: A research library and benchmark for federated machine learning},
  author={He, Chaoyang and Li, Songze and So, Jinhyun and Zeng, Xiao and Zhang, Mi and Wang, Hongyi and Wang, Xiaoyang and Vepakomma, Praneeth and Singh, Abhishek and Qiu, Hang and others},
  journal={arXiv preprint arXiv:2007.13518},
  year={2020}
}

@article{fan2023generative,
  title={Generative diffusion models on graphs: Methods and applications},
  author={Fan, Wenqi and Liu, Chengyi and Liu, Yunqing and Li, Jiatong and Li, Hang and Liu, Hui and Tang, Jiliang and Li, Qing},
  journal={arXiv preprint arXiv:2302.02591},
  year={2023}
}

@book{hamilton2020graph,
  title={Graph representation learning},
  author={Hamilton, William L},
  year={2020},
  publisher={Morgan \& Claypool Publishers}
}
\section*{Appendix}
\appendix

\section{Details of Datasets}
\label{appx:data}

\begin{table*}[t!]
    \centering
    \begin{tabular}{|c|c|c|c|c|c|c|c|c|}
    \hline
         \textbf{Dataset} &\textbf{Domain}& \textbf{\# Graphs} & \textbf{Avg. \# Nodes}  & \textbf{Avg. \# Edges} & \textbf{Avg. \# Degrees} & \textbf{\# Node features} &\textbf{\# Classes}\\\hline
         \textbf{MUTAG}&Molecules&188&17.93&19.79&2.11&7&2\\
         \textbf{ENZYMES}&Proteins&600&32.60&124.3&7.63&3&6\\
         \textbf{PROTEINS}&Proteins&1,113&39.06&72.82&3.73&3&2\\
         \textbf{IMDB-BINARY}&Social network&1,000&19.77&96.53&9.77&135&2\\
         \textbf{IMDB-MULTI}&Social network&1,500&13.00&65.94&10.14&88&3\\\hline
    \end{tabular}
    \caption{Statistics of datasets}
    \label{tab:data}
   \vspace{-0.2in}
\end{table*}

\begin{table*}[t!]
    \centering
    \scalebox{0.92}{
    \begin{tabular}{|c|c|c|c|c|c|c|c|c|}
    \hline
         \multirow{4}{*}{\textbf{Settings}}&\multicolumn{5}{c|}{\textbf{Single dataset}}&\multicolumn{2}{c|}{\textbf{Across-dataset}}&{\textbf{Across-domain}}\\\cline{2-9}
         & &&&&& Protein &Social &Molecular\&Protein\&Social\\\cline{7-9}
         & MUTAG&PROTEINS&ENZYMES&IMDB-B&IMDB-M&PROTEINS&IMDB-B&MUTAG\&ENZYMES\\
         & &&&&&\&ENZYMES&\&IMDB-M&\&IMDB-B\&IMDB-M\\\hline
         \textbf{Avg. Struc. hetero.}&0.16($\pm 0.02$)&0.28($\pm 0.06$)&0.28($\pm 0.05$)&0.27($\pm 0.05$)&0.28($\pm 0.05$)&0.37($\pm 0.13$)&0.28($\pm 0.22$)&0.39($\pm 0.20$)\\\hline
         \textbf{Avg. Feat. hetero.}&0.13($\pm 0.09$)&0.13($\pm 0.10$)&0.16($\pm 0.12$)&0.18($\pm 0.12$)&0.19($\pm 0.15$)&0.15($\pm 0.11$)&0.15($\pm 0.13$)&0.21($\pm 0.13$)\\\hline
    \end{tabular}
}
   \vspace{-0.05in}
   \caption{ Heterogeneity for different settings between different clients.} \label{tab:heterogeneity}
   \vspace{-0.2in}
\end{table*}

\begin{table*}
    \centering
    \scalebox{0.93}{
    \begin{tabular}{|c|c|c|c|c|c|c|c|c|c|c|c|c|c|c|c|c|c|c|c|c|}
    \hline
    \multirow{3}{*}{\textbf{Settings}}
    &\multicolumn{8}{c|}{MUTAG dataset}&\multicolumn{8}{c|}{PROTEINS dataset}\\\cline{2-17}
    &\multicolumn{4}{c|}{k=1}&\multicolumn{4}{c|}{k=2}&\multicolumn{4}{c|}{k=1}&\multicolumn{4}{c|}{k=2}\\\cline{2-17}
    &Deg.$\downarrow$&Clus.$\downarrow$&Orb.$\downarrow$&Avg.$\downarrow$&Deg.$\downarrow$&Clus.$\downarrow$&Orb.$\downarrow$&Avg.$\downarrow$&Deg.$\downarrow$&Clus.$\downarrow$&Orb.$\downarrow$&Avg.$\downarrow$&Deg.$\downarrow$&Clus.$\downarrow$&Orb.$\downarrow$&Avg.$\downarrow$\\\hline
         \textbf{w/o}&0.032&0.122&0.004&0.052&0.004&0.131&0.001&0.045&0.137&0.704&0.443&0.428&0.276&0.387&0.147&0.270\\\hline
         \textbf{with}&0.047&0.103&0.019&0.057&0.032&0.040&0.030&0.034&0.179&0.645&0.427&0.417&0.257&0.324&0.257&0.279\\\hline
    \end{tabular}
}
   \caption{\label{tab:quality_diffusion} Quality of the generative diffusion models with or without the graph label channel. Deg., Clus., and Orb. represent the MMD values in terms of degree distribution, clustering coefficients, and the number of occurrences of orbits with 4 nodes, respectively, and Avg. denotes the average MMD value across Deg., Clus., and Orb.} 
   \vspace{-0.2in}
\end{table*}

\begin{figure*}
    \centering
    \begin{subfigure}[b]{.32\textwidth}
      \centering
     \includegraphics[width=\textwidth]{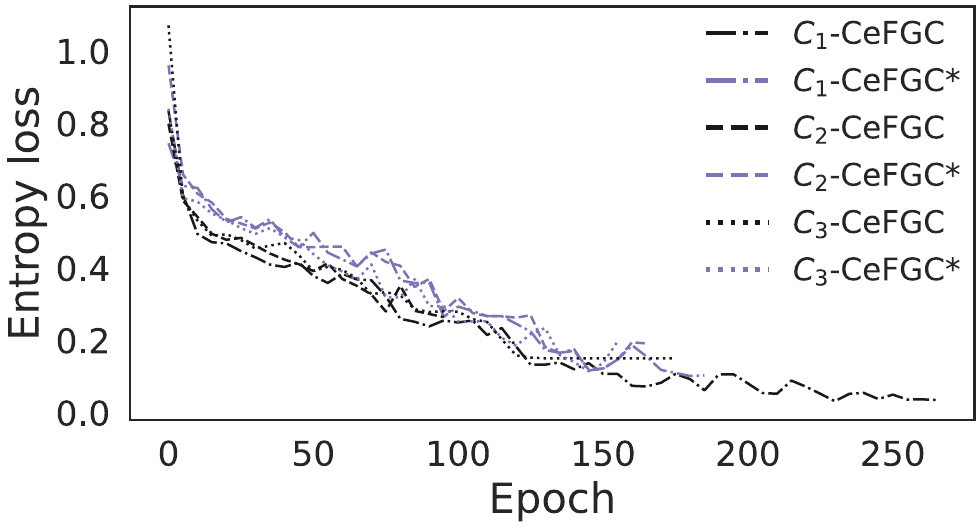}
     \vspace{-0.2in}
    \caption{\label{fig:loss-single-bi} Single-dataset Setting}
    \end{subfigure}
    \begin{subfigure}[b]{.32\textwidth}
      \centering
     \includegraphics[width=\textwidth]{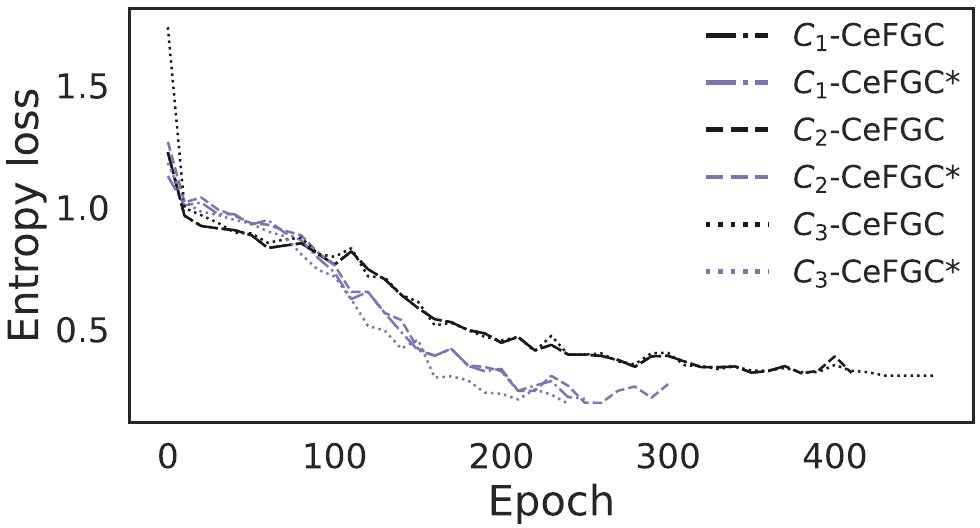}
     \vspace{-0.2in}
    \caption{\label{fig:loss-social-bi} Across-dataset Setting}
    \end{subfigure}
    \begin{subfigure}[b]{.32\textwidth}
      \centering
    \includegraphics[width=\textwidth]{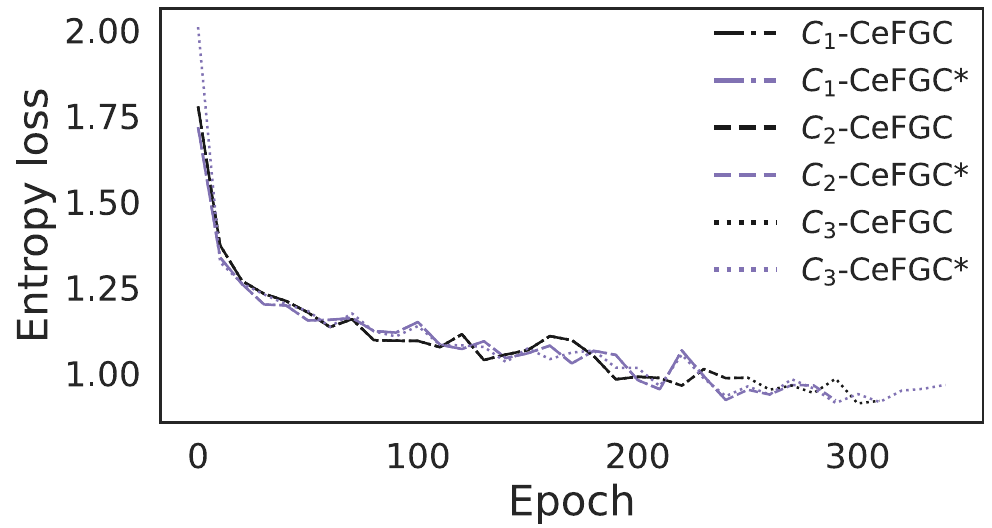}
    \vspace{-0.2in}
    \caption{\label{fig:loss-mix} Across-domain Setting}
    \end{subfigure}
    \vspace{-0.1in}
\caption{\label{fig:loss-local-more} Convergence analysis of local GNN models through the learning curve (IMDB-B dataset).} 
\vspace{-0.05in}
\end{figure*}

{\bf Statistics of datasets.} We use five real-world datasets for graph classification in our paper, namely MUTAG, ENZYMES, PROTEINS, IMDB-BINARY, and IMDB-MULTI,\footnote{All the datasets are downloaded from torch\_geometric package. For IMDB-BINARY and IMDB-MULTI datasets that do not have node features, we follow \cite{xie2021federated} and use one-hot degree features as the node features.} which serve as benchmarks for evaluating graph-based models in various domains \cite{morris2020tudataset}. 
Table \ref{tab:data} provides details of these datasets.  

{\bf Non-IID distribution.} We measure the average heterogeneity of features and structures for each setting and include the results in Table \ref{tab:heterogeneity}. For structure heterogeneity, first, we train the graph embedding for each graph with Anonymous Walk Embeddings (AWEs) \cite{ivanov2018anonymous, xie2021federated}. Second, for each pair of clients, we calculate the Jensen-Shannon (JS) distance between the AWEs of each graph pair across the two clients. Third, we calculate the structure heterogeneity as the average JS distance across all graph pairs for all client pairs. For feature heterogeneity, we first calculate the feature similarity of the linked edges within each graph with cosine similarity. Second, for each pair of clients, we compute the JS distance between the feature similarity of each graph pair across the two clients. Finally, we assess feature heterogeneity as the average JS distance across all graph pairs for all client pairs. From Table \ref{tab:heterogeneity}, we observe significant heterogeneity in all three settings (i.e., the data follows the non-IID distribution in the three settings). Furthermore, the structure heterogeneity increases from the single-dataset setting to across-datasets setting, and across-domains setting. For example, the structure heterogeneity values for PROTEINS and ENZYMES are 0.28 in the single-dataset setting, but increase to 0.37 in the across-dataset setting and 0.39 in the across-domain setting.

\vspace{-0.1in}
\section{ Performance of the Generative Diffusion Models with and without the Graph Label Channel}
\label{sec:results_diffusion}
By following \cite{zhang2023survey, niu2020permutation, chen2022nvdiff, huang2022graphgdp, jo2022score, luo2022fast}, we assess the quality of the diffusion model by using maximum mean discrepancy (MMD) to compare the distributions of graph statistics, such as the degree, the clustering coefficient, and the number of occurrences of orbits with 4 nodes between the same number of generated and test graphs. As shown in Table \ref{tab:quality_diffusion}, we observe the comparable diffusion model performance with and without the graph label channel, demonstrating that the model retains its quality with the inclusion of the graph label channel.

\section{Convergence of Local GNN Models under Three Settings}
\label{appx:loss curve}
Figure \ref{fig:loss-local-more} exhibits the learning curves of local GNNs under the single-dataset, across-dataset, and across-domain settings on IMDB-B. We observe that all local GNNs are well-converged within 500 epochs.

\newpage
\end{document}